\documentclass[runningheads]{llncs}

\usepackage{eccv}

\usepackage{eccvabbrv}

\usepackage{graphicx}
\usepackage{booktabs}
\usepackage{multirow}
\usepackage{makecell}

\usepackage{wrapfig}
\usepackage{colortbl}

\usepackage[accsupp]{axessibility}  % Improves PDF readability for those with disabilities.

\usepackage{hyperref}

\usepackage{orcidlink}

\begin{document}

% ---------------------------------------------------------------
% MuRF table(rebuttal table.1) 넣기-(6 view / 10 view main table에 넣기) (LIRF도 넣을지말지 고민 중) 
% finetune table(rebuttal table.2 llff에서만) 넣기
% pixelsplat table(re10k/acid) 넣기
% Train, finetune, rendering time, GPU memory supp table 넣기
% intro에  theoretical innovation / Reasons for GNeRF research 강조하기
% supp corres figure2,3 넣기
% supp table 1 & main table 2 합치기
% video supple에 추가하기

% \title{Enhancing Generalizable Neural Rendering via Geometry-Driven Multi-Reference Texture Transfer} 
\title{GMT: Enhancing Generalizable Neural Rendering via Geometry-Driven Multi-Reference Texture Transfer} 

\titlerunning{Geometry-Driven Multi-Reference Texture Transfer}

% \author{Youngho Yoon\inst{1}\orcidlink{0000-1111-2222-3333} \and
% Hyun-Kurl Jang\inst{2,3}\orcidlink{1111-2222-3333-4444}}


% Springer Heidelberg, Tiergartenstr.~17, 69121 Heidelberg, Germany
\newcommand\CoAuthorMark{\footnotemark[\arabic{footnote}]}

\author{Youngho Yoon\orcidlink{0009-0003-4346-8260}\thanks{Equal contribution.}, Hyun-Kurl Jang\orcidlink{0009-0003-7943-3326}\protect\CoAuthorMark, and Kuk-Jin Yoon\orcidlink{0000-0002-1634-2756}}

\authorrunning{Yoon et al.}
\institute{Visual Intelligence Lab., KAIST \\
% \email{lncs@springer.com}
\email{\{dudgh1732, jhg0001, kjyoon\}@kaist.ac.kr} \\
\url{https://github.com/yh-yoon/GMT}
}

\maketitle

\begin{abstract}

Novel view synthesis (NVS) aims to generate images at arbitrary viewpoints using multi-view images, and recent insights from neural radiance fields (NeRF) have contributed to remarkable improvements. 
% Recently, generalizable NeRFs (G-NeRFs) research eliminates scene-specific optimization which is an inevitable process of NeRFs and constructs the radiance field on-the-fly, thereby accelerating the NVS speed to make it available for real-world applications.
Recently, studies on generalizable NeRF (G-NeRF) have addressed the challenge of per-scene optimization in NeRFs. The construction of radiance fields on-the-fly in G-NeRF simplifies the NVS process, making it well-suited for real-world applications.
% Recently, generalizable NeRF(G-NeRF) studies have addressed the challenge of per-scene optimization in NeRFs. 
% Construction of radiance fields on-the-fly in G-NeRF simplifies NVS process, making it well-suited for real-world applications.
% Meanwhile, G-NeRF still struggles in representing fine details for NVS effectively due to the absence of optimization process.
Meanwhile, G-NeRF still struggles in representing fine details for a specific scene due to the absence of per-scene optimization, even with texture-rich multi-view source inputs.
As a remedy, we propose a \textbf{G}eometry-driven \textbf{M}ulti-reference \textbf{T}exture transfer network (GMT) available as a plug-and-play module designed for G-NeRF. 
 Specifically, we propose ray-imposed deformable convolution (RayDCN), which aligns input and reference features reflecting scene geometry. 
Additionally, the proposed texture preserving transformer (TPFormer) aggregates multi-view source features while preserving texture information.
Consequently, our module enables direct interaction between adjacent pixels during the image enhancement process, which is deficient in G-NeRF models with an independent rendering process per pixel. 
This addresses constraints that hinder the ability to capture high-frequency details.
Experiments show that our plug-and-play module consistently improves G-NeRF models on various benchmark datasets.
\end{abstract}

\section{Introduction}
\label{sec:intro}

\begin{figure}[t]
    \centering
    \includegraphics[width=0.75\linewidth]{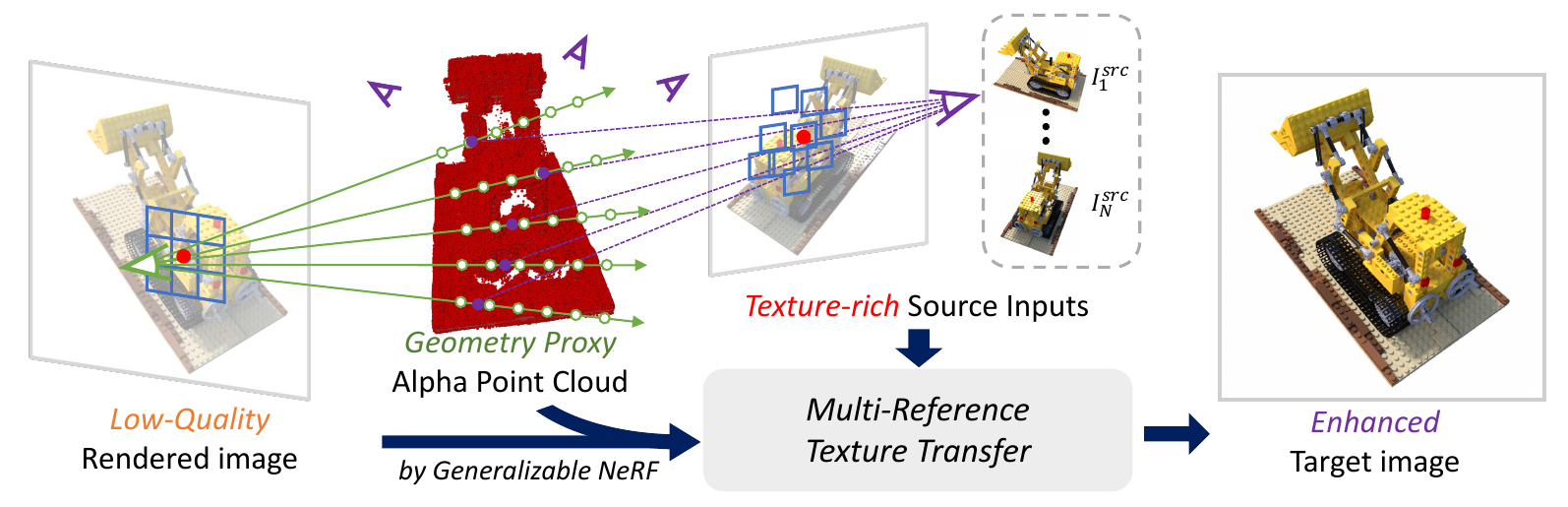}
    % \vspace{-7pt}
    \caption{The proposed Geometry-driven Multi-reference Texture transfer (GMT) model.}
    % \vspace{-20pt}
    \label{fig:intro}
\end{figure}
Novel view synthesis (NVS) is an approach that synthesizes an image of an arbitrary viewpoint from multi-view source images. Early studies on NVS have primarily explored image-based rendering~\cite{levoy2023light,gortler2023lumigraph,yao2018mvsnet,shum2000review}.
Recently, neural radiance fields~\cite{mildenhall2021nerf}  (NeRF) proposed a volume rendering method through 5D radiance field optimization that extracts densities and colors from 5D inputs (3D locations and 2D directions). 
NeRF-based approach has inspired numerous studies in NVS tasks, resulting in remarkable performance enhancements.

Recent studies have pointed out that NeRF requires per-scene optimization for NVS and proposed various generalizable NeRF (G-NeRF) methods~\cite{trevithick2020grf,yu2021pixelnerf,wang2021ibrnet, suhail2022generalizable, wang2022attention, xu2023murf}.
These studies enable NVS on-the-fly without per-scene optimization through cross-scene generalization. Since then, several studies have attempted to improve rendering quality and speed based on G-NeRF. Neuray~\cite{liu2022neural} and GeoNeRF~\cite{ohari2022geonerf} improve understanding of scene geometry and occlusion by utilizing multi-view stereo methods. Additionally, recent research leveraging 3D Gaussian splatting~\cite{kerbl20233d} in generalizable neural rendering has significantly accelerated rendering speed~\cite{charatan2023pixelsplat}.
Despite these advances, generalizable NVS approaches commonly encounter difficulties in accurately representing high-frequency details. The methods proposed in G-NeRF studies often struggle to capture the local textures for a specific scene due to the absence of per-scene optimization, even with texture-rich multi-view source inputs.

To solve this problem, as shown in Fig.~\ref{fig:intro}, we aim to enhance rendered images by 1) geometric priors derived from the rendering process of G-NeRF and 2) transferring high-frequency details of texture-rich source images. Inspired by prior studies in reference-based image enhancement~\cite{zhang2019image,jiang2021robust,zhang2023lmr}, we develop a network that improves NVS performance by transferring textures acquired from multi-view source images onto rendered images from G-NeRF. Enhancement is done in a plug-and-play manner, demonstrating improved results within seconds.

To achieve the goal, we propose a ray-imposed deformable convolution network, RayDCN, that conducts geometry-driven reference feature alignment with the target view feature map.
RayDCN determines the spatial location on source images for deformable convolution by leveraging the alpha values of sampled points obtained during the G-NeRF rendering process. Through this approach, we make deformable convolution incorporate the scene geometry by utilizing the estimated spatial location of source views of each ray. While utilizing estimated geometry for source-target feature alignment, we additionally employ correspondence matching to handle occlusion and inaccuracies in geometry estimation.

We also introduce a texture-preserving transformer, called TPFormer, designed for multi-reference feature aggregation. TPFormer transfers the texture from multi-reference images to the target view image through two steps. Firstly, view-dependent attention performs self-attention with input features of view differences and the corresponding view to obtain preliminary information for reference-texture selection. Secondly, reference-texture selection process aggregates features from multiple reference images via feature selection, considering the relationships between multiple references obtained during the view-dependent attention step. 
TPFormer seamlessly integrates multi-view source features extracted by RayDCN without impairing texture information.

Consequently, our model handles multi-ray features with a receptive field on the target viewpoint, while G-NeRF models handle single-ray. G-NeRF models estimate the color of each pixel independently in a batch-wise manner, which causes a deficiency in direct interaction among adjacent pixels in the target image. This deficiency hampers the model from representing subtle variations or intricate patterns in the image. However, our module induces interactions among adjacent rays and generates a superb NVS result.
We experimentally demonstrate that our method consistently enhances the performance of existing G-NeRF models on various NVS benchmark datasets. In summary, our contributions are as follows: 
% \vspace{-5pt}
\begin{itemize}%[leftmargin=*]
   \item We introduce a Geometry-driven Multi-reference Texture transfer network (GMT) for generalizable neural rendering.
   \item We propose ray-imposed deformable convolution, which performs feature alignment reflecting scene geometry.
   \item We propose a texture-preserving transformer for source features aggregation with preserving texture features.
   \item Our plug-and-play module consistently improves generalizable NeRF model performances on various benchmark datasets without additional training.
\end{itemize}

% \vspace{-15pt}
\section{Related Works}
% \vspace{-5pt}
\subsection{Image based rendering}
Image-based rendering (IBR) methods render novel views directly from input images without the necessity of 3D scene representations. Early research in IBR~\cite{levoy2023light,shum2000review,gortler2023lumigraph} rendered novel views from dense input image sets using the 4D plenoptic function. Studies adopting geometry proxies~\cite{yao2018mvsnet,chaurasia2013depth,debevec1998efficient,thies2019deferred,huang2020adversarial,penner2017soft,gortler2023lumigraph} have demonstrated that satisfactory rendering quality could be attained using depth or mesh data obtained from input images.
% While these approaches reduced sampling density, their performance depended on the accuracy of the geometry proxy. 
As deep learning has progressed, numerous  works utilizing deep neural networks have come to the forefront~\cite{choi2019extreme,hedman2018deep,kalantari2016learning,riegler2020free,thies2018ignor,xu2019deep}. LFNR~\cite{suhail2022light} has leveraged the strength of classic IBR technique~\cite{levoy2023light} which is resistant to reflections. NeRF-based models like IBRNet~\cite{wang2021ibrnet} and Neuray~\cite{liu2022neural} applied volume rendering techniques akin to NeRF using features from input images and geometric information. 
 
 % Methods such as IBRNet~\cite{wang2021ibrnet} and Neuray~\cite{liu2022neural} have emerged, which use sparse input images to obtain deep features, depth, and apply volume rendering techniques similar to NeRF in research.
% \vspace{-10pt}
\subsection{Generalizable NeRF}
Synthesizing photo-realistic images has been a long-standing area of research interest. 
% One effective and impressive solution to address view synthesis is using 
Neural scene representations, exemplified by the Neural Radiance Fields (NeRF)~\cite{mildenhall2021nerf} is effective and impressive solution for view synthesis.
% NeRF optimizes continuous scenes through a 5D neural radiance field, followed by volume rendering. 
% NeRF has disentangled processes of modeling and rendering. 
Following works of NeRF improved rendering quality~\cite{roessle2022dense,wei2021NeRFingmvs,deng2022depth} and optimization, and rendering speed~\cite{sun2022direct, chen2022tensorf,fridovich2022plenoxels,yu2021plenoctrees,muller2022instant,xu2022point}. However, NeRF retains the limitation that per-scene optimization is required to perform novel view synthesis. Various works have explored generalizable NeRF (G-NeRF)~\cite{wang2021ibrnet,cao2022fwd,ohari2022geonerf,liu2022neural,yang2023contraNeRF,wang2022attention,suhail2022generalizable,chen2021mvsnerf,yu2021pixelnerf} to overcome this limitation. G-NeRF learns a view interpolation function from source images, enabling cross-scene generalization. In G-NeRF, the typical approach involves using volume rendering to aggregate information obtained from images, such as deep features, depth maps, and cost volumes~\cite{wang2021ibrnet,liu2022neural,ohari2022geonerf,chen2021mvsnerf,xu2023murf}.
GNT~\cite{wang2022attention} and GPNR~\cite{suhail2022generalizable} leverage transformers as feature aggregators to enhance the interaction of information within a single ray, resulting in the direct acquisition of RGB values for each pixel. PixelSplat~\cite{charatan2023pixelsplat} introduces generalizable volume rendering using scene parameterization based on 3D Gaussian primitives~\cite{kerbl20233d}.
Despite the advance, G-NeRF still maintains the independence of the rendering process for each pixel, leading to the failure of transferring fine textures in source images. To solve this problem, we propose a novel method to integrate information from reference images using multi-ray aggregation. Furthermore, our approach applies to various G-NeRF models and collectively enhances their performance.
% \vspace{-10pt}
\subsection{Reference-based Image Enhancement}
Image enhancement tasks aim to rectify degradation or elevate overall visual quality of given image. Some studies also incorporate reference images to retain the fine textures and intricate details found within reference images~\cite{zou2023reference,zhang2019image,zheng2018crossnet,liu2023reference,li2022reference}. In the context of reference-based image super-resolution~\cite{zhang2019image,yang2020learning,zheng2018crossnet,zhang2022rrsr,xia2022coarse,jiang2021robust,lu2021masa,zhang2023lmr,cao2022reference}, the approach involves transferring additional details from high-resolution reference images to low-resolution input images. The common practice in reference-based super-resolution is to utilize a single reference image, but some approaches employ multiple reference images~\cite{zhang2023lmr,pesavento2021attention,yoon2023cross}.
In reference-based deblurring tasks~\cite{zou2023reference,liu2023reference,li2022reference}, high-quality features extracted from reference images enhance the quality of blurred input images. In reference-based restoration tasks, it is typical to establish image-based correspondences between input images and reference images to identify applicable reference features.
The present study introduces a novel approach that leverages geometric priors to estimate correspondences while retaining multi-view consistency. The correspondences, derived from sampled points and the associated alpha values generated during the rendering process, are jointly utilized with the image pair correlation to enhance the accuracy of locating relevant features.

\section{Method}
\label{Method}

\begin{figure*}[t]
    \centering
    \includegraphics[width=0.9\linewidth]{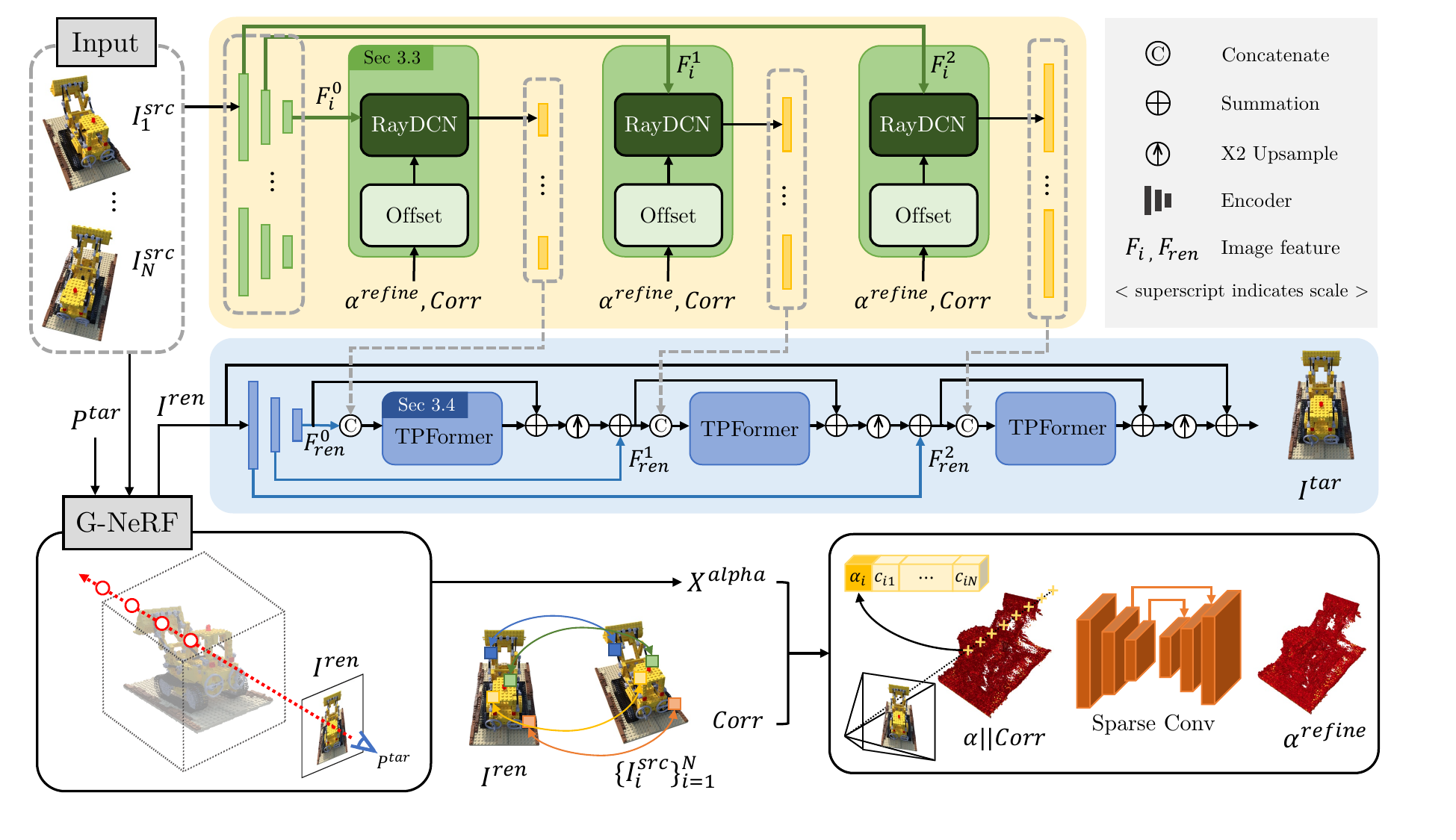}
    % \vspace{-17pt}
    \caption{Overall framework of the Geometry-driven Multi-reference Texture transfer (GMT) model. When generalizable NeRFs (G-NeRF) renders novel view image $I^{ren}$ with N source images $\{I^{src}_i\}_{i=1}^N$ and a target camera pose $P^{tar}$, the process inherently generates alpha point cloud $X^{alpha}$ for volume rendering process. Using $\alpha^{refine}$ extracted from $\alpha$ and correlation values $Corr$, RayDCN enables feature alignment considering scene geometry.
    % through $\alpha^{refine}$ and $Corr$. 
    Subsequently, TPFormer conducts multi-reference feature aggregation and the model generates final output $I^{tar}$.}
    % \caption{tbd.}
    % \vspace{-15pt}
    \label{fig:framework}
\end{figure*}

\subsection{Preliminary} 
In common, Generalizable Neural Radiance Fields (G-NeRF) generate a novel view image $I^{ren}$ from $N$ source-view images $\{I^{src}_i\}_{i=1}^N$ taken from sparse views, without the process of scene optimization performed in NeRF.
A ray $p(r)$ with direction vector $r$ emitted from the target view camera center $o$ can be expressed as $p(r) = o +zr$. 
After sampling K points among the points existing on ray $p(r)$, alpha $\{\alpha_i\}_{i=1}^K$ and colors $\{c_i\}_{i=1}^K$ of each point are estimated.
In this process, G-NeRF learns a network that aggregates source features projected from $\{I^{src}_i\}_{i=1}^N$ to estimate $\alpha_i$ and $c_i$ without per-scene optimization. 
% Afterwards, the final rendered color is calculated using the following equation:
Finally, the color $c$ of ray $p(r)$ is calculated using the following equation:
\begin{equation}
    c =  \prod_{i=1}^{K} h_i  c_i =\prod_{i=1}^{K}T_i  \alpha_i  c_i
  \label{eq:volumerender}
\end{equation}
where  the hitting probability $h_i=T_i \alpha_i$ and the accumulated transmittance $T_i = \prod_{j=1}^{i-1} (1-\alpha_j)\alpha_i$.

\subsection{Multi-Reference Texture Transfer Network}

% We propose a geometry-driven multi-reference texture transfer network to overcome the methodological limitations of existing studies of G-NeRF that perform independent ray sampling and color rendering. Inspired by the potential for texture transfer from precise local textures and high-frequency details in the source images $\{I^{src}_i\}_{i=1}^N$, we utilize these images as reference images for the image enhancement model. It finally derives an enhanced novel-view synthesis result $I^{tar}$. As shown in Fig.~\ref{fig:framework}, we formulate our proposed approach through the following stages.

% We present a geometry-driven multi-reference texture transfer network aimed at overcoming the methodological constraints observed in previous studies of G-NeRF, particularly those characterized by independent ray sampling and color rendering. Inspired by the potential for texture transfer from precise local textures and high-frequency details in the source images $\{I^{src}_i\}_{i=1}^N$, we employ these images as reference inputs for our image enhancement model. This strategic utilization serves as a primary motivation for our proposed method, culminating in the generation of an enhanced novel-view synthesis result denoted as $I^{tar}$. As in Fig.~\ref{fig:framework}, we formulate our proposed approach through the following stages.
We present a geometry-driven multi-reference texture transfer network aiming to overcome the methodological constraints observed in previous studies of G-NeRF—particularly those characterized by independent ray sampling and color rendering for each pixel. 
Inspired by the potential for texture transfer from unblemished local textures and high-frequency details in the source images $\{I^{src}_i\}_{i=1}^N$, we leverage these images for our image enhancement model. 
Our model addresses the common texture scarcity issue and blurry artifacts in G-NeRF by leveraging valuable information in source images, ultimately extracting improved rendering results denoted as $I^{tar}$. As in Fig.~\ref{fig:framework}, we formulate our proposed approach through the following stages.

\noindent\textbf{Rendering by G-NeRF.} 
%우리는 proxy scene geometry 및 texture transfer network의 initial image로 사용되는 query image를 얻기 위해 가장 먼저 $\{I^{src}_i\}_{i=1}^N$로부터 G-NeRF를 통해 novel view synthesis를 수행한다.
We initially conduct a novel view synthesis from $\{I^{src}_i\}_{i=1}^N$ using one of the G-NeRF models to generate the rendered image $I^{ren}$, which will be used as a query image of our model. The neural rendering process inherently generates alpha point cloud $X^{alpha}$, which includes alpha values from the volume rendering process. We utilize $X^{alpha}$ as proxy geometry and the initial input for the following offset estimation network.

\noindent\textbf{Feature Extraction.} 
%우리는 source image로부터 multi-scale local texture features를 얻기 위해 먼저 VGG model을 이용하여 feature extraction을 수행한다. multi-scale features를 사용함으로써 high-level의 semantic, 그리고 low-level의 textual features를 수행할 수 있도록 한다.
We initiate feature extraction with the VGG model to derive multi-scale texture features $\{ \mathbf{F}_i \}_{i=1}^N$ from input source images $\{I^{src}_i\}_{i=1}^N$ and $\mathbf{F}_{ren}$ from rendered image.

\noindent\textbf{Offset Estimation and Feature Alignment.} 
%우리는 $I_{ren}$을 query image로 사용하여 i번째 source feature $\mathbf{F}_i$를 target viewpoint에 대응하는 feature map으로 swap해오기 위한 stage를 수행한다. 본 stage에서, 우리는 ray-imposed deformable convolution network (RayDCN)을 제안한다(Sec.~\ref{RayDCN}). RayDCN은 $I_{ren}$과 $X^{alpha}$를 활용하여, multi-view constraint를 고려한 offset estimation을 수행한다. 이후, deformable convolution을 통해서 feature swapping을 수행한다. 
% In this stage, we conduct feature alignment which find correspondence between source and target feature and .
In this stage,  we perform feature alignment by finding correspondence between the source and target features. This correspondence position features beneficial to the target region at their corresponding coordinates. To achieve this, we propose a ray-imposed deformable convolution network (RayDCN) (Sec.~\ref{RayDCN}).  RayDCN leverages refined alpha and correlation for offset estimation while considering multi-view constraints. Subsequently, feature alignment is executed through deformable convolution.

\noindent\textbf{Reference Feature Aggregation.} 
%swap된 source features를 이용하여, 우리는 $I^{ren}$으로부터 extract된 rendered feature와의 융합을 위한 reference feature aggregation을 수행한다. 본 과정에서, 우리는 referece image의 local textures 보존이 가능한 aggregation module인 texture-preserving traniTPFormer를 제안하고(Sec.~\ref{TPFormer}), 해당 모듈을 사용하여 multi-scale feature aggregation을 수행한다. 총 세 번의 upscaling을 통해, target viewpoint에 대한 최종 result인 $I^{tar}를 합성한다.$
Using the aligned features, we conduct reference feature aggregation to merge them with the rendered features extracted from $I^{ren}$. In this process, we introduce texture-preserving transformer (TPFormer), an aggregation module designed to maintain the local textures of the reference images (Sec.~\ref{TPFormer}). We synthesize the final result for the target viewpoint image $I^{tar}$ by fusion and upscaling three times.

% \subsection{Geometry-aided Deformable Convolution (GDC)}
% \vspace{-10pt}
\subsection{Ray-Imposed Deformable Convolution (RayDCN)}
\label{RayDCN}

\begin{figure}[t!]
    \centering
    \includegraphics[width=0.65\linewidth]{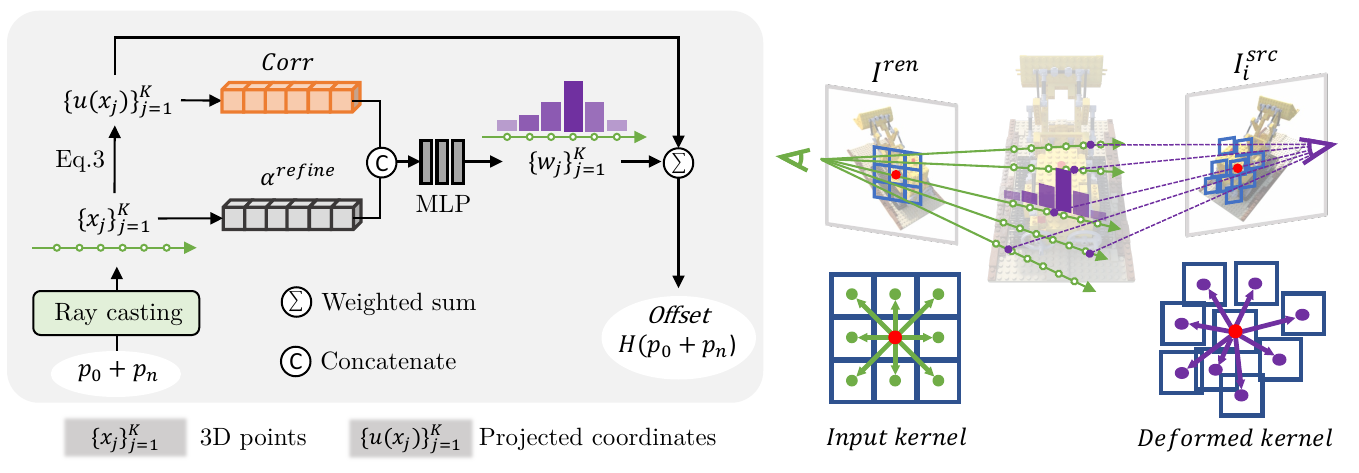  }
    % \vspace{-4pt}
    \caption{Ray-imposed Deformable Convolution (RayDCN). It has a deformed kernel shape considering scene geometry and aggregates the source features of multiple rays.}
    % \vspace{-10pt}
    \label{fig:convolution}
\end{figure}

The general framework of multi-reference-based enhancement (MRE) first involves finding the corresponding points between images through optical flow estimation~\cite{zheng2018crossnet} or correspondence matching~\cite{jiang2021robust,zhang2023lmr}.
In MRE, source images are considered referenceable, but in most cases, reference images are not obtained from scenes identical to the target image. 
In contrast, in G-NeRF, the source and target images are always obtained from the same scene. Hence, multi-view geometry is applicable while enhancing the rendered image $I^{ren}$ from G-NeRF. In light of this circumstance, we propose RayDCN for multi-reference texture transfer which finds corresponding points utilizing $\{I^{src}_i\}_{i=1}^N$ and $I^{ren}$ obtained through G-NeRF, and $X^{alpha}$ for proxy representation of scene geometry.
Raw alpha values in G-NeRF include unreliable noise due to inaccurately estimated geometry. Thus, we employed inter-image correlation to obtain a reliable and refined alpha. we first generate point cloud features of $N+1$ channels by concatenating the alpha values of sampled points and the correlation values $Corr$ between $I^{ren}$ and $\{I^{src}_i\}_{i=1}^N$. Generated point cloud feature is processed through sparse convolution, MinkowskiEngine~\cite{choy20194d}, and returns refined alpha value $\alpha^{refine}$ of the points.
% considering multi-view geometry. RayDCN 
Following this, as depicted in Fig.~\ref{fig:convolution}, we estimate offset which decides reference point in source image $I^{src}_i$ of view $i$.
This estimation is performed based on the $Corr$ and $\alpha^{refine}$. 
When single 2D coordinate $p$ on $I^{ren}$ is decided, we can specify the points $\{x_j\}_{j=1}^K$ on the ray passing through $p$. To estimate the offset of $p$, we calculate weight $\{w_j\}_{j=1}^K$ from $\{x_j\}_{j=1}^K$ and its allotted $\alpha^{refine}$ and $Corr$. $\{w_j\}_{j=1}^K$ is obtained through the following equation.
\begin{equation}
  \{w_j\}_{j=1}^K = \mathbf{S}(MLP(M_{val} \odot \{\alpha^{refine} || Corr\})
)
  \label{eq:sigmaj}
\end{equation}
where $M_{val}$ is a valid projection mask that determines whether 3D point $x_j$ can be projected onto $I^{src}_i$ and $\mathbf{S}$ indicates softmax function.
% and the weight of each point is $\{w_j\}_{j=1}^K$. 
Then, $u(x_j)$ is the projected coordinate on the image plane of $I^{src}_i$ corresponding to $x_j$. 
To compute $u(x_j)$, the equation is as follows:
\begin{equation}
  u(x_j)=K^{s}_i (R^{r}_i D_j (K^{t})^{-1} x_j+T^{r}_i)
  \label{eq:sigmaj}
\end{equation}
where $K^{s}_i$ and $K^{t}$ are the intrinsic matrices of $I^{src}_i$ and $I^{tar}$, respectively. $R^{r}_i$ and $T^{r}_i$ are the relative rotation and relative translation between $I^{src}_i$ and $I^{tar}$. $D_j$ is the depth of $x_j$.
The obtained $w_j$ and $u(x_j)$ are combined through a weighted summation to generate $\mathbf{H}(p)$ which is the offset of $p$.
% with $u(x_j)$ to obtain the reference coordinate $\mathbf{H}(p_0)$.
$\mathbf{H}(p)$ is calculated by this equation:
\begin{equation}
  \mathbf{H}(p) = \{\sum_{j} w_j \cdot u(x_j) / \sum_{j}w_j\} - p
  \label{eq:sigmaj}
\end{equation}
where $w_j$ is used as a weight, and $u(x_j)$ is the value. Because $p+\mathbf{H}(p)$ is a affine combination with the coordinates $\{u(x_j)\}_{j=1}^K$ as an element, it always exists on the epipolar line.
Therefore, we can narrow the candidate of offsets through epipolar constraints. Calculated $p+\mathbf{H}(p)$ is the reference point of $p$ derived from a singe-ray. 
% Naive convolution then goes through a convolution filter with neighboring coordinates on the source view. 
% However,
Then, RayDCN performs feature alignment of the source image features to the target viewpoint.
The introduced RayDCN deforms the convolution kernel shape, reflecting scene geometry and aggregating multi-ray derived features. As shown in Fig.~\ref{fig:convolution}, the offsets $\{\mathbf{H}(p_0+p_n)\}_\mathcal{R}$ of rays passing through $p_0$ and neighboring pixels on target view will be goes through convolution filter:
% Finally, the equation of RayDCN is as follows:
% \vspace{-4pt}
\begin{equation}
  \mathbf{y}(p_0) = \sum_{p_n \in \mathcal{R}}\mathbf{w}(p_n)\cdot \mathbf{F}(p_0+p_n+\mathbf{H}(p_0+p_n))
  % \vspace{-4pt}
  \label{eq:dconv}
\end{equation}
where $\mathbf{y}(p_0)$ is the value of $p_0$ on the output feature map $\mathbf{y}$, $\mathbf{w}$ is weights of convolution filter, and $\mathbf{F}$ is the input feature map. $\mathcal{R}$ is regular grids of the convolution filter and defined as,
\begin{equation}
  \mathcal{R} = \{(-1,-1),(-1,0),...,(0,1),(1,1) \}
  \label{eq:dconv}
\end{equation}

\subsection{Texture-Preserving Transformer (TPFormer)}
\label{TPFormer}

\begin{figure}[t!]
    \centering
    \includegraphics[width=0.8\linewidth]{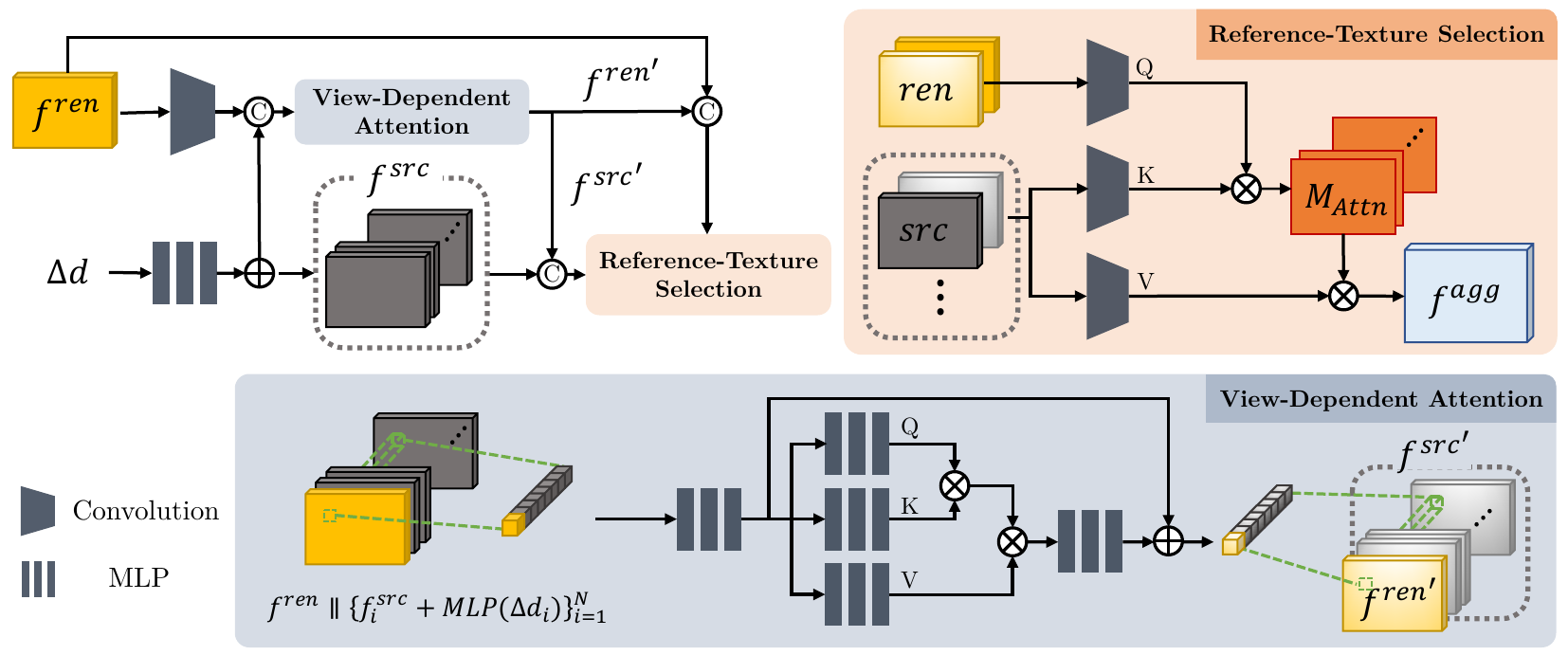}
    % \vspace{-6pt}
    \caption{Texture-Preserving Transformer (TPFormer). TPFormer aggregates features from multiple source views while preserving textures from the source image.}
    \label{fig:aggregation}
    % \vspace{-16pt}
\end{figure}

The architectural configurations commonly employed in G-NeRF models involve aggregating input features considering ray-direction. Notably, IBRNet~\cite{wang2021ibrnet} and Neuray~\cite{liu2022neural} utilize a multi-layer perceptron (MLP), whereas  GeoNeRF~\cite{ohari2022geonerf} and GNT~\cite{wang2022attention} incorporate  a multi-head attention (MHA) layer.
However, we found that the MLP and MHA structures tend to blend detailed texture features derived from input images, leading to blurry artifacts.
As illustrated in Fig.~\ref{fig:aggregation}, we introduce a novel component termed the texture-preserving transformer (TPFormer) to address this issue. 
TPFormer aims to perform appropriate aggregation according to the ray-direction of the source view image features while preserving their high-frequency details.
TPFormer includes a view-dependent attention (VA) module and a reference-texture selection (RS) module. 
% The VA module performs pixel-wise self-attention on the features of $I^{ren}$ and $I^{src}_i$. 
Before operating TPFormer, RayDCN conduct alignment and generate aligned source image feature $f_i^{src}$, which has a texture information of $I^{src}_i$. 
% using a pair of  $I^{ren}$ and $I^{src}_i$ as an input.
% Each 2D coordinate in target view calculates feature value $\mathbf{y}(p_0)$ from corresponding point pixel in source view. 
% The output feature map of RayDCN, $f_i^{src}$, which has $I^{src}_i$ as an input, will be  $f_i^{src}$ and aligned with target view feature. Aligned feature map of $N$ source views will be generated independently. 
% By this process, we obtain the i-th aligned target view feature $f_i^{src}$, which has a texture information of $I^{src}_i$. 
% We generate aligned features of $N$ source views $\{f_i^{src}\}_{i=1}^{N}$ independently.
After that, we conduct the VA module in TPFormer to perform pixel-wise self-attention on the features of $I^{ren}$ and $I^{src}_i$. 
$f_i^{src}$ is added by relative view direction embedding and combined with the target view feature before self-attention is applied. This process aggregates source view features based on relative viewing direction, as described by the following equation:
\begin{equation}
    f^{ren}{'}, \{f_i^{src}{'}\}_i = Net_{attn}(f^{ren}\parallel \{f^{src}_i+Net_{mlp}(\Delta d^{src}_i)\}_i)
    \label{eq:va}
\end{equation}
% \begin{equation}
%     \begin{aligned}
%         f^{ren}{'},& \{f_i^{src}{'}\}_{i=1}^{N}  \\
%         &=Attn(f^{ren}\parallel \{f^{src}_i+MLP(\Delta d_i)\}_{i=1}^{N}) 
%     \end{aligned}
% \end{equation}
where $\Delta d_i$ is view direction of i-th source view relative to the target view. The RS module receives input features $f^{ren}{''}$,$f^{src}{''}$, which is a fusion of resulting features of VA and aligned features from RayDCN.
\begin{equation}
    f^{ren}{''}= f^{ren} \;\parallel\; f^{ren}{'}
    \label{eq:va}
    % \vspace{-16pt}
\end{equation}
\begin{equation}
    f_i^{src}{''} = f^{src}_i \;\parallel\; f_i^{src}{'}
    \label{eq:va}
    % \vspace{-10pt}
\end{equation}
% a concatenate of the features generated from the VA module and the features $f^{src}$ of $\{I^{src}_i\}$. 
%그 다음, 우리는 RS module을 통해 f^{ren}{''}와 유사한 f_i^{src}{''} 영역을 attention을 이용하여 찾고, 해당 attention map을 통해 feature aggregation을 수행한다.
%RS 모델에서는 각각의 source view에서 얻어진 fren''들을 aggregate한다. 먼저 fren''을 통해 얻어진 query 와 fsrc로 부터 생성된 key를 사용한 pixel-wise attention을 i-th view의 attention map을 구한다. 구해진 attention map 과 source view feature의 aggregation feature를 통해 TPFormer의 결과인 fagg를 얻는다. 
RS module aggregates $N$ features from source view $\{f_i^{src}{''}\}_{i=1}^{N}$ with $f_i^{ren}{''}$. The attention map for i-th view is calculated using the query from $f^{ren}{''}$ and key from $f_i^{src}{''}$. Final result of TPFormer $f^{agg}$ is derived from aggregation of attention map and value from $f_i^{src}{''}$, equation is as follows:    
% Then, we calculate attention map $M_{Attn}$ to find $f_i^{src}{''}$ regions similar to $f^{ren}{''}$ through RS module, and perform feature aggregation through the corresponding attention map. In RS module, $I^{ren}$ performs pixel-wise attention with $f^{src}{''}$. Since $f^{src}$ is used as input, texture information can be maintained. To compute aggregated feature $f^{agg}$, the equation is as follows:
% \vspace{-8pt}
\begin{equation}
    f^{agg} = \sum_{i}^{N}(Attn(Q(f^{ren}{''}),K(f_i^{src}{''})) \cdot V(f_i^{src}{''}))
    \label{eq:va}
    % \vspace{-4pt}
\end{equation}
% \subsection{Multi-Refernce Enhancement}
% TPFormer는 VA를 통해 얻을 수 있는 각 feature와 viewing direction 사이의 관계를 사용한 feature aggregation을 진행하기 때문에 reference view에 존재하는 texture를 target view로 적절히 transfer 할 수 있다.
TPFormer performs feature aggregation using the relationship between each feature and viewing direction obtained through VA. Therefore, textures in the reference view can be seamlessly transferred to the target view image.

\section{Experiments}
% \vspace{-6pt}
\subsection{Implementation Details}
% \vspace{-6pt}
\noindent\textbf{Training Datasets.}
%training dataset 설명
% 우리의 training dataset은 Neuray를 학습하기 위해 사용된 dataset 중 일부가 사용되었다. 그 중에는 synthetic dataset과 real dataset들이 포함되어 있다. synthetic dataset으로는 Google Scanned Objects, real dataset으로는 dtu dataset, LLFF dataset, spaces dataset을 사용했다. Google Scanned Objects은 총 1023개의 object centered image로 구성되어 있으며 본 논문에서는 각각의 object에 포함된 이미지들 중 10장의 이미지만을 사용하였다. dtu dataset은 4개의 scene을 제외한 나머지 scene 109개를 train set으로 사용하였다. forward-facing dataset으로는 LLFF 중 35개의 scenes, space dataset 89개의 scene을 train 과정에서 사용하였다.
% Google Scanned Objects / Spaces / DTU dataset / LLFF dataset / 
Our training datasets consist of synthetic and real datasets. For synthetic data, we use Google Scanned Objects dataset~\cite{francis2022google}, which contains 1023 objects, and we employ 10 images from each object. For real datasets, we used 109 scenes from the DTU dataset~\cite{jensen2014large}, 35 scenes from the Real Forward-Facing dataset~\cite{mildenhall2019local} and 89 scenes from the Spaces dataset~\cite{flynn2019deepview}. We use a black background for Google Scanned Objects and DTU dataset.

\noindent\textbf{Test Datasets.}
%Test dataset 설명
%evaluation은 DTU dataset, Synthetic NeRF, LLFF Forward-Facing 으로 구성되어 있다. DTU dataset 중 trainset과 중복되지 않는 4 scene(birds,tools, bricks, and snowman)을 사용하였다. LLFF dataset과 Synthetic NeRF는 8개의 scene으로 구성되었다. 각 dataset의 evaluation resolution은 DTU dataset 800 x 600, LLFF dataset 1008 x 756, Synthetic NeRF은 800 x 800이다. DTU dataset과 Synthetic NeRF의 경우 black-background로 evaluation 진행했다. 
In evaluation, we use DTU dataset~\cite{jensen2014large}, Synthetic NeRF~\cite{mildenhall2021nerf}, and Real Forward-Facing~\cite{mildenhall2019local}. Within the DTU dataset, we use four scenes (birds, tools, bricks, and snowman) at 800$\times$600 resolution. Real Forward-Facing dataset and Synthetic NeRF dataset are both consist of 8 scenes. The evaluation resolutions are 1008$\times$756 for Real Forward-Facing dataset, and 800$\times$800 for Synthetic NeRF dataset. DTU dataset and Synthetic NeRF are evaluated with a black background.

\begin{table}[t]
\centering
\caption{Novel view synthesis results on DTU, Synthetic NeRF, and Real Forward-Facing datasets.
% Through our model, a significant and consistent improvement is observed across IBRNet, GNT, GeoNeRF, Neuray.
Ours consistently exhibits a  significant improvement across IBRNet, GNT, GeoNeRF, Neuray, and MuRF.
\textbf{Bold} indicates the best, and \underline{underline} indicates the second best results.}
% \vspace{-8pt}
\resizebox{0.9\textwidth}{!}{%
\setlength{\tabcolsep}{3pt}
\begin{tabular}{c||c|c|c||c|c|c||c|c|c}
\Xhline{3\arrayrulewidth}

\multirow{2}{*}{Method}       & \multicolumn{3}{c||}{DTU}    &   \multicolumn{3}{c||}{Synthetic NeRF}   &      \multicolumn{3}{c}{Real Forward-Facing}  \\
\cline{2-10}
             & PSNR($\uparrow$)    & SSIM($\uparrow$)    & LPIPS($\downarrow$)  & PSNR($\uparrow$)    & SSIM($\uparrow$)    & LPIPS($\downarrow$)  & PSNR($\uparrow$)    & SSIM($\uparrow$)    & LPIPS($\downarrow$)  \\
\Xhline{3\arrayrulewidth}
pixelNeRF~\cite{yu2021pixelnerf}         & 19.40   & 0.463  & 0.447  & 22.65  & 0.808  & 0.202  & 18.66  & 0.588   & 0.463  \\
\hline
% GPNR~\cite{suhail2022generalizable}         & \underline{28.5}   & \textbf{0.932}  & 0.167  & 26.48  & \textbf{0.944}  & 0.091  & 25.72  & \textbf{0.880}   & 0.175  \\
% \hline
IBRNet~\cite{wang2021ibrnet}       & 26.76 & 0.879  & 0.136  & 25.03 & 0.900    & 0.102  & 25.19  & 0.822  & 0.173  \\
\hline
MVSNeRF~\cite{chen2021mvsnerf}      & 23.83  & 0.723  & 0.286  & 25.15  & 0.853  & 0.159  & 21.18  & 0.691  & 0.301  \\
\hline
GNT~\cite{wang2022attention}          & 25.46 & 0.818 & 0.171 & 23.11 & 0.763 & 0.141  & 25.54 & 0.835  & 0.177 \\
\hline
GeoNeRF~\cite{ohari2022geonerf}      & \multicolumn{3}{c||}{-}   & \underline{29.59}  & 0.933  & 0.071  & 25.64  & 0.847  & 0.150   \\
\hline
% GPNR~\cite{suhail2022generalizable}         & \underline{28.5}   & \textbf{0.932}  & 0.167  & 26.48  & \textbf{0.944}  & 0.091  & 25.72  & \textbf{0.880}   & 0.175  \\
% \hline
ContraNeRF~\cite{yang2023contraNeRF}   & 27.69  & 0.904  & 0.129  & 27.92  & 0.930   & \underline{0.060}   & 25.44  & 0.842  & 0.178  \\
\hline
Neuray~\cite{liu2022neural}       & \underline{28.37} & \underline{0.906} & \underline{0.112} & 28.36 & 0.928  & 0.071 & 25.43 & 0.833 & 0.161 \\
\hline
MuRF~\cite{xu2023murf}       & 24.87 & 0.870 & 0.183 & 22.26 & 0.612  & 0.225 & \underline{26.49} & \underline{0.909} & 0.143 \\
\Xhline{3\arrayrulewidth}
\multirow{2}{*}{Ours+IBRNet~\cite{wang2021ibrnet}}  & 26.96      & 0.893      & 0.124      & 25.29      & 0.909      & 0.089      & 25.57      & 0.839      & 0.154     \\
 & \cellcolor[HTML]{FFE9CE} (0.20 \textcolor{red}{$\uparrow$})    & \cellcolor[HTML]{FFE9CE} (0.014 \textcolor{red}{$\uparrow$})    & \cellcolor[HTML]{FFE9CE} (0.012 \textcolor{red}{$\downarrow$})  & \cellcolor[HTML]{FFE9CE} (0.26 \textcolor{red}{$\uparrow$})    & \cellcolor[HTML]{FFE9CE} (0.009 \textcolor{red}{$\uparrow$})    & \cellcolor[HTML]{FFE9CE} (0.013 \textcolor{red}{$\downarrow$})  & \cellcolor[HTML]{FFE9CE} (0.38 \textcolor{red}{$\uparrow$})    & \cellcolor[HTML]{FFE9CE} (0.017 \textcolor{red}{$\uparrow$})    & \cellcolor[HTML]{FFE9CE} (0.019 \textcolor{red}{$\downarrow$}) \\
\hline
\multirow{2}{*}{Ours+GNT~\cite{wang2022attention}}  & 25.60      & 0.832      & 0.155      & 23.23      & 0.773      & 0.124      & 25.81      & 0.850     & 0.157      \\
 & \cellcolor[HTML]{FFE9CE} (0.14 \textcolor{red}{$\uparrow$})    & \cellcolor[HTML]{FFE9CE} (0.014 \textcolor{red}{$\uparrow$})    & \cellcolor[HTML]{FFE9CE} (0.016 \textcolor{red}{$\downarrow$})  & \cellcolor[HTML]{FFE9CE} (0.12 \textcolor{red}{$\uparrow$})    & \cellcolor[HTML]{FFE9CE} (0.010 \textcolor{red}{$\uparrow$})    & \cellcolor[HTML]{FFE9CE} (0.017 \textcolor{red}{$\downarrow$})  & \cellcolor[HTML]{FFE9CE} (0.27 \textcolor{red}{$\uparrow$})    & \cellcolor[HTML]{FFE9CE} (0.015 \textcolor{red}{$\uparrow$})    & \cellcolor[HTML]{FFE9CE} (0.020 \textcolor{red}{$\downarrow$}) \\
\hline
\multirow{2}{*}{Ours+GeoNeRF~\cite{ohari2022geonerf}} & \multicolumn{3}{c||}{\multirow{2}{*}{-}}     & \textbf{29.81}      & \textbf{0.942}      & \textbf{0.052}      & 25.80      & 0.857      & \textbf{0.137}     \\
 & \multicolumn{3}{c||}{}  & \cellcolor[HTML]{FFE9CE} (0.22 \textcolor{red}{$\uparrow$})    & \cellcolor[HTML]{FFE9CE} (0.009 \textcolor{red}{$\uparrow$})    & \cellcolor[HTML]{FFE9CE} (0.019 \textcolor{red}{$\downarrow$})  & \cellcolor[HTML]{FFE9CE} (0.16 \textcolor{red}{$\uparrow$})    & \cellcolor[HTML]{FFE9CE} (0.010 \textcolor{red}{$\uparrow$})    & \cellcolor[HTML]{FFE9CE} (0.013 \textcolor{red}{$\downarrow$}) \\
\hline
\multirow{2}{*}{Ours+Neuray~\cite{liu2022neural}}     & \textbf{28.72}      & \textbf{0.920}      & \textbf{0.101}      & 28.96      & \underline{0.936}      & 0.062      & 25.82      & 0.850      & 0.142 \\
 & \cellcolor[HTML]{FFE9CE} (0.35 \textcolor{red}{$\uparrow$})    & \cellcolor[HTML]{FFE9CE} (0.014 \textcolor{red}{$\uparrow$})    & \cellcolor[HTML]{FFE9CE} (0.011 \textcolor{red}{$\downarrow$})  & \cellcolor[HTML]{FFE9CE} (0.60 \textcolor{red}{$\uparrow$})    & \cellcolor[HTML]{FFE9CE} (0.008 \textcolor{red}{$\uparrow$})    & \cellcolor[HTML]{FFE9CE} (0.010 \textcolor{red}{$\downarrow$})  & \cellcolor[HTML]{FFE9CE} (0.39 \textcolor{red}{$\uparrow$})    & \cellcolor[HTML]{FFE9CE} (0.017 \textcolor{red}{$\uparrow$})    & \cellcolor[HTML]{FFE9CE} (0.019 \textcolor{red}{$\downarrow$}) \\
 \hline
\multirow{2}{*}{Ours+MuRF~\cite{xu2023murf}}     & 24.91      & 0.872      & 0.180     & 22.36      & 0.614      & 0.217      & \textbf{26.81}      & \textbf{0.915}      & \underline{0.139} \\
 & \cellcolor[HTML]{FFE9CE}(0.04 \textcolor{red}{$\uparrow$})    & \cellcolor[HTML]{FFE9CE}(0.002 \textcolor{red}{$\uparrow$})    & \cellcolor[HTML]{FFE9CE}(0.003 \textcolor{red}{$\downarrow$})  & \cellcolor[HTML]{FFE9CE}(0.10 \textcolor{red}{$\uparrow$})    & \cellcolor[HTML]{FFE9CE}(0.002 \textcolor{red}{$\uparrow$})    & \cellcolor[HTML]{FFE9CE}(0.008 \textcolor{red}{$\downarrow$})  & \cellcolor[HTML]{FFE9CE}(0.32 \textcolor{red}{$\uparrow$})    & \cellcolor[HTML]{FFE9CE}(0.016 \textcolor{red}{$\uparrow$})    &\cellcolor[HTML]{FFE9CE} (0.004 \textcolor{red}{$\downarrow$}) \\
\Xhline{3\arrayrulewidth}
\end{tabular}
}
% \vspace{-3pt}
\label{table:result_table}%
\end{table}

\begin{table}[t]
\centering
\caption{Novel view synthesis results on ACID and RealEstate10k datasets.
\textbf{Bold} indicates the best, and \underline{underline} indicates the second best results.}
% \vspace{-8pt}
\resizebox{0.6\textwidth}{!}{%
\setlength{\tabcolsep}{3pt}
\begin{tabular}{c||c|c|c||c|c|c}
\Xhline{3\arrayrulewidth}

\multirow{2}{*}{Method}       & \multicolumn{3}{c||}{ACID}    &   \multicolumn{3}{c}{RealEstate10k}    \\
\cline{2-7}
             & PSNR($\uparrow$)    & SSIM($\uparrow$)    & LPIPS($\downarrow$)  & PSNR($\uparrow$)    & SSIM($\uparrow$)    & LPIPS($\downarrow$)  \\
\Xhline{3\arrayrulewidth}
pixelNeRF~\cite{yu2021pixelnerf}          & 20.97 & 0.547 & 0.533 & 20.43 & 0.589 & 0.550   \\
\hline
GPNR~\cite{suhail2022generalizable}     & 25.28 & 0.764 & 0.332   & 24.11  & 0.793  & 0.255     \\
\hline
Du et al.~\cite{du2023learning}   & 26.88  & 0.799  & 0.218  & 24.78  & 0.820   & 0.213  \\
\hline
pixelSplat~\cite{charatan2023pixelsplat}   &   \underline{28.10} & \underline{0.846} & \underline{0.122} & \underline{25.86} & \underline{0.865} & \underline{0.110}  \\
\Xhline{3\arrayrulewidth}
\multirow{2}{*}{Ours+pixelSplat~\cite{charatan2023pixelsplat}}      & \textbf{28.87} & \textbf{0.855} &\textbf{0.107}&\textbf{26.40} & \textbf{0.871}  & \textbf{0.101}\\
 &\cellcolor[HTML]{FFE9CE} (0.77 \textcolor{red}{$\uparrow$})    &\cellcolor[HTML]{FFE9CE} (0.009 \textcolor{red}{$\uparrow$})    &\cellcolor[HTML]{FFE9CE} (0.015 \textcolor{red}{$\downarrow$})  &\cellcolor[HTML]{FFE9CE} (0.54 \textcolor{red}{$\uparrow$})    &\cellcolor[HTML]{FFE9CE} (0.006 \textcolor{red}{$\uparrow$})    &\cellcolor[HTML]{FFE9CE} (0.009 \textcolor{red}{$\downarrow$})  \\
\Xhline{3\arrayrulewidth}
\end{tabular}
}
% \vspace{-13pt}
\label{table:pixelsplat}%
\end{table}
\begin{figure*}[t!]
    % \vspace{-5pt}
    \centering
    \includegraphics[width=0.86\linewidth]{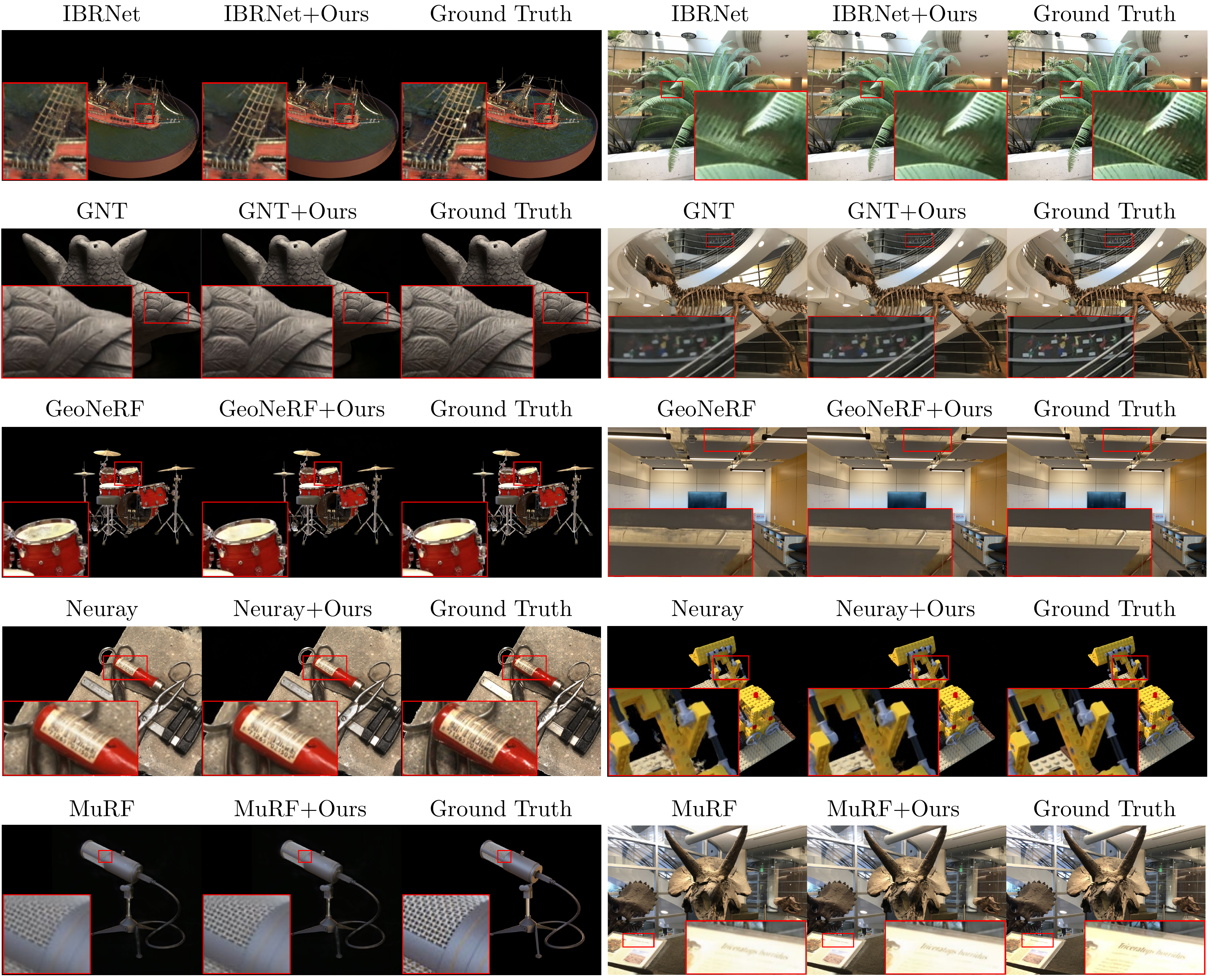}
    % \vspace{-7pt}
    \caption{Qualitative comparisons of generalizable NeRF models on DTU, Real Forward-Facing, and Synthetic NeRF datasets.} 
    \label{fig:mainvisual1}
\end{figure*}
\begin{figure*}[t!]
    % \vspace{-17pt}
    \centering
    \includegraphics[width=0.86\linewidth]{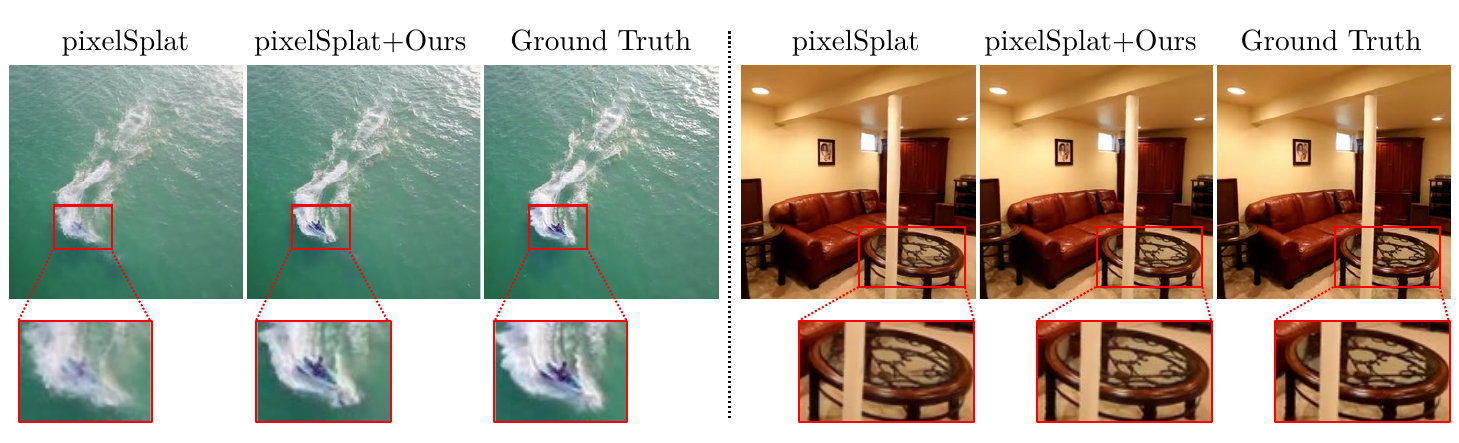}
    % \vspace{-5pt}
    \caption{Qualitative comparisons of pixelSplat results on ACID(left) and RealEstate10k(right) datasets.} 
    % \vspace{-25pt}
    \label{fig:mainvisual}
\end{figure*}

\noindent\textbf{Network Details.}
%본 모델에서는 offset estimation을 진행할 때 reference based SR에서 사용하는 correspondence matching과 geometry-driven 방식을 융합하여 성능 향상을 이루었다. 이때 C2-matching의 단순히 rendered 이미지 feature와 reference 이미지 feature를 사용하는  correspondence matching module을 사용했다. Image feature를 얻기 위해서는 texture를 잘 represent하는 것으로 알려진 vgg extractor~\cite{simonyan2014very}를 사용했다r~\cite{gatys2015texture,gatys2016image}. 또한 multi-scale feature를 활용하기 위해 vgg extractor의 relu1 1, relu2 1,and relu3 1을 encoder 처럼 사용하였다. g nerf에서 발생한 alpha 값을 process 할 때 이는 sparse한 3d points이기 때문에 MinkowskiNet [14],specifically MinkUNet14을 사용하였다.
% In this model, we achieved improvement in offset estimation by combining image based correspondence matching, as used in reference-based Super-Resolution (SR), with a geometry-driven approach. 
Offset estimation for the proposed RayDCN integrates an image-based correspondence matching module with a geometry-driven approach. The correspondence matching module is based on C2-Matching~\cite{jiang2021robust}, which only uses features from rendered and reference images. We use the VGG extractorr~\cite{simonyan2014very} to extract image features, which is known for its ability to represent texture~\cite{gatys2015texture,gatys2016image}. Additionally, to leverage multi-scale features, $\textnormal{relu1\_1}$, $\textnormal{relu2\_1}$, and $\textnormal{relu3\_1}$ layers in VGG extractor are used as encoders. When processing the alpha values generated by G-NeRF, which are sparse 3D points, we utilized MinkowskiNet~\cite{choy20194d}, specifically MinkUNet14.

\noindent\textbf{Dataset Generation.}
% 우리의 모델은 g nerf model의 rendered image를 enhance 해주는 모델로 novel view Image 와 해당 rendering 과정에서 생성되는 alpha 값을 input을 Input으로 받는다. 본 논문에서 training을 진행할 때에는 학습 과정을 가속화하기 위해 g nerf에서 train dataset에 해당하는 이미지들을 rendering 하고 해당 이미지와 alpha 값들을 3d point 위치와 함께 저장해 놓고 사용하였다. training set을 구성하는데 사용된 g nerf 모델은 실험 세팅과 동일한 train, test dataset split을 가지고 있는 SOTA model 중 하나인 Neuray를 사용하였다. 우리 모델을 학습하는데 사용되는 reference 이미지는 g nerf 모델에서 사용된 source 이미지들을 사용하여 그 외의 추가적인 이미지 사용은 없었다. alpha value 값은 각 ray 별로 $h_i을 기준으로 sorting 하였고 그 중 값이 큰 12개의 sampled points만을 사용하였다.
Our model is an enhancer that improves the performance of G-NeRF models. It takes novel view images and the corresponding alpha values generated during the rendering process as input. In order to accelerate the training and evaluation process, we pre-rendered novel view images from the G-NeRF model. 
We stored these images, their corresponding alpha values, and 3D point positions. 
The alpha values were sorted for each ray based on the $h_i$, and we selected the top 12 sampled points and corresponding alpha. For constructing the training dataset, we utilized Neuray~\cite{liu2022neural}, which follows the same train-test dataset split as our experimental setup and one of the state-of-the-art (SOTA) models. 
% Our model relied on the source images employed by the G-NeRF model for reference, without incorporating additional images.

%g-nerf중 neuray preprocessing으로 training dataset 생성
%network model architecture 구체적인 설명
%training hyper-parameter 설명

% \vspace{-10pt}
\noindent
\begin{minipage}[t!]{0.56\textwidth}%
% \vspace{-5pt}
\centering
\captionof{table}{Novel view synthesis with per-scene finetuning on Real Forward-Facing dataset.}
\label{table:ft_llff}

\resizebox{0.99\textwidth}{!}{%
\setlength{\tabcolsep}{10pt}
\begin{tabular}{c||c|c|c}
\Xhline{3\arrayrulewidth}

{Method}      &  PSNR($\uparrow$)    & SSIM($\uparrow$)    & LPIPS($\downarrow$)  \\
% \multirow{2}{*}{Method}      &      \multicolumn{3}{c}{Real Forward-Facing}  \\
% \cline{2-4}
%               & PSNR($\uparrow$)    & SSIM($\uparrow$)    & LPIPS($\downarrow$)  \\
\Xhline{3\arrayrulewidth}
% InstantNGP        & 24.57  & 0.798  & 0.211  \\
% \hline
% Tri-MipRF       & 26.83  & 0.845  & 0.192 \\
LLFF~\cite{mildenhall2019local}        & 23.93  & 0.798  & 0.212  \\
\hline
NeRF~\cite{mildenhall2021nerf}       & 26.36  & 0.811  & 0.250 \\
\hline
NeX~\cite{wizadwongsa2021nex}       & 27.03  & 0.890  & 0.182  \\
\hline
NLF~\cite{suhail2022light}       & 28.03  & 0.917  & 0.129 \\
\Xhline{3\arrayrulewidth}
Neuray-ft~\cite{liu2022neural}        & 27.40  & 0.869  & 0.129  \\
\hline
GNT-ft~\cite{wang2022attention}      & \underline{30.73}  & \underline{0.943}  & \textbf{0.081} \\
\Xhline{3\arrayrulewidth}
% Ours+     & 27.56 & (0.16 \textcolor{red}{$\uparrow$})     & 0.876 \\ {Neuray-ft} & (0.007 & hiu\textcolor{red}{$\uparrow$})    & 0.125   \\
% \hline
% Ours+     & 27.56 & (0.16 \textcolor{red}{$\uparrow$})     & 0.876 \\ {Neuray-ft} & (0.007 & hiu\textcolor{red}{$\uparrow$})    & 0.125   \\
% \Xhline{3\arrayrulewidth}
\multirow{1}{*}{Ours+Neuray-ft~\cite{liu2022neural}}     & 27.56\: (0.16 \textcolor{red}{$\uparrow$})     & 0.876 \: (0.007 \textcolor{red}{$\uparrow$})    & 0.125\:  (0.004 \textcolor{red}{$\downarrow$})   \\
 % &  (0.16 \textcolor{red}{$\uparrow$})    & (0.007 \textcolor{red}{$\uparrow$})    & (0.004 \textcolor{red}{$\downarrow$}) \\
\hline
\multirow{1}{*}{Ours+GNT-ft~\cite{wang2022attention}}  & \textbf{31.03} \:(0.30 \textcolor{red}{$\uparrow$})     & \textbf{0.946} \:(0.003 \textcolor{red}{$\uparrow$})      & \underline{0.082}\:  (0.001 \textcolor{red}{$\uparrow$})    \\
 % &     & (0.003 \textcolor{red}{$\uparrow$})    & (0.001 \textcolor{red}{$\uparrow$}) \\
\Xhline{3\arrayrulewidth}
\end{tabular}
}
% \vspace{10pt}
\end{minipage}%
\;
\begin{minipage}[t!]{0.43\textwidth}%
% \vspace{-12pt}
\centering
\captionof{table}{Quantitative comparisons of reference-based image enhancement models.}
\label{table:enhance_speed}%

\resizebox{0.99\textwidth}{!}{%
\setlength{\tabcolsep}{3pt}
\begin{tabular}{c||c||c|c|c||c|c} 
\Xhline{3\arrayrulewidth}
\multirow{2}{*}{Model}                & \multirow{2}{*}{\begin{tabular}[c]{@{}c@{}}
Ref.\\Type\end{tabular}} & \multicolumn{3}{c||}{Speed (sec)}                  & \multirow{2}{*}{\begin{tabular}[c]{@{}c@{}}Params\\(M)\end{tabular}} & \multirow{2}{*}{\# Ref.}  \\ 
\cline{3-5}
                                      &                           & DTU  & Syn. & RFF. &                                                                      &                          \\ 
\Xhline{3\arrayrulewidth}
C2-Matching~\cite{jiang2021robust} & Single                    & 0.32 & 1.22           & 0.49                & 8.9                                                                  & 1                        \\ 
\Xhline{3\arrayrulewidth}
MRefSR~\cite{zhang2023lmr}      & \multirow{3}{*}{Multi} & 1.99 & 8.98           & 3.25                & \underline{23.7}                                                                 & \textbf{8}                        \\ 
\cline{1-1}\cline{3-7}
NeRFLiX~\cite{zhou2023nerflix}    &                           & \underline{1.48} & \underline{2.35}           & \underline{1.95}                & 35.2                                                                 & 2                        \\ 
\cline{1-1}\cline{3-7}
Ours        &                           & \textbf{1.07} & \textbf{1.68}           & \textbf{1.61}                & \textbf{9.1}                                                                  & \textbf{8}                        \\
\Xhline{3\arrayrulewidth}
\end{tabular}
}
\end{minipage}%

\subsection{Comparison with Generalizable NeRFs}

% \vspace{-2pt}

\noindent\textbf{Experimental Setup.}
%main comparison 
%우리는 state-of-the-art NeRF generalization methods과의 비교 실험을 진행한다. 그 models는 IBRNet, MVSNeRF, Neuray, GeoNeRF, GPNR, ContraNeRF, GNT을 포함한다. 방법론에서 설명했듯이, 우리는 초기에 G-NeRF base model로부터 target-view에 대한 initial rendered image와 alpha point cloud를 얻는다. 따라서, 우리는 base model로써 IBRNet, GNT, GeoNeRF, Neuray를 사용했을때에 대한 기존 모델들과의 성능비교를 수행한다. test dataset은 총 세 개로, DTU, NeRF Synthetic, LLFF datasets를 사용한다. Evaluation metrics는 PSNR, SSIM LPIPS를 사용하며, source view는 총 8개의 이미지를 사용한다. Base models는 개별적으로 학습된 pretrained model을 사용한다. 또한, GeoNeRF는 training dataset에 DTU test dataset의 일부가 포함되어 있기 때문에, 본 model에 대한 DTU의 성능평가는 제외된다.
We conduct comparative experiments with state-of-the-art generalizable NeRF methods. The models include IBRNet~\cite{wang2021ibrnet}, MVSNeRF~\cite{chen2021mvsnerf}, Neuray~\cite{liu2022neural}, GeoNeRF~\cite{ohari2022geonerf}, GPNR~\cite{suhail2022generalizable}, ContraNeRF~\cite{yang2023contraNeRF}, GNT~\cite{wang2022attention}, and MuRF~\cite{xu2023murf}. 
Our model aims to enhance the rendering performance of the G-NeRF models. Therefore, we selected baseline models for comparison, and evaluated the enhanced results in comparison to the performance of the baseline models. IBRNet, GNT, GeoNeRF, Neuray, and MuRF are chosen as the baseline model for this purpose. The evaluation is conducted using a total of three datasets: DTU~\cite{jensen2014large}, Synthetic NeRF~\cite{mildenhall2021nerf}, and Real Forward-Facing~\cite{mildenhall2019local}. Evaluation metrics include PSNR, SSIM~\cite{hore2010image}, and LPIPS~\cite{zhang2018unreasonable}. Additionally, input images that are used for NVS are utilized as reference images for enhancement purposes. Base models use individually learned pretrained models. Pretrained GeoNeRF was trained on a dataset that includes part of the DTU test dataset; therefore, GeoNeRF is excluded from the evaluation of the DTU dataset.

\noindent\textbf{Quantitative Analysis.}
%Table
% Table~\ref{table:result_table}에서 볼 수 있듯이,  우리의 모델들은 모든 metrics, 모든 세 개의 데이터셋들(DTU, Synthetic NeRF, 그리고 Real Forward-Facing datasets)에 대해서 consistent한 성능향상을 이끈다. DTU dataset에서, ours(+Neuray)는 Neuray보다 PSNR(high is better)이 0.3426 더 높으며 비교모델들 중 가장 높은 수치를 가진다. Ours(+GNT)는 GNT보다 LPIPS(low is better)가 0.016 더 낮아서 해당 metric 기준으로 가장 높은 성능향상을 가진다. Synthetic NeRF dataset에서, ours(+GeoNeRF)는 GeoNeRF보다 PSNR(high is better)이 0.2273 더 높으며 비교모델들 중 가장 높은 수치를 가진다. Ours(+GNT)는 GNT보다 LPIPS(low is better)가 0.017 더 낮아서 해당 metric 기준으로 가장 높은 성능향상을 가진다. Real forward facing dataset에서는, ours(+Neuray)는 Neuray보다 PSNR(high is better)이 0.3826 더 높으며 비교모델들 중 가장 높은 수치를 가진다. 이와 비슷하게, ours(+IBRNet) 또한 IBRNet보다 PSNR이 0.3802 더 높은 성능향상을 가진다. 본 결과를 토대로, 우리 모델은 NeRF generalization setting의 대부분의 SOTA 모델들에 대해 합리적이고 consistent한 성능향상을 일으킨다고 결론지을 수 있다.
As shown in Table~\ref{table:result_table}, our models lead to consistent improvements across all metrics and all datasets.
% (DTU, Synthetic NeRF, and Real Forward-Facing datasets). 
In the DTU dataset, performance improved up to 0.35 from the baseline model.
% has a 0.35 higher PSNR than Neuray and is the highest-valued model. 
%Ours(+GNT) gets LPIPS (low is better) 0.016 lower than GNT, so it has the highest performance improvement based on that metric. 
In the Synthetic NeRF dataset, performance improved up to 0.6 from the baseline model.
% In the Synthetic NeRF dataset, ours(+GeoNeRF) has a 0.22 higher PSNR than GeoNeRF and is the highest-valued model. 
%Ours(+GNT) has LPIPS (low is better) 0.017 lower than GNT, so it has the highest performance improvement based on that metric. 
In the real forward-facing dataset, performance improved up to 0.39 from the baseline model. %Similarly, ours(+IBRNet) also has a performance improvement of 0.3802 higher in PSNR than IBRNet. Based on these results, we can conclude that our model produces reasonable and consistent performance improvements for most SOTA models in generalizable neural rendering.

\noindent\textbf{Qualitative Comparison and Analysis.}
% Fig.~\ref{fig:mainvisual}에서 보이듯이, 기존 IBRNet, GeoNeRF, GNT, Neuray와 비교했을 때, our models는 더 선명한 texture를 가지고 있으며, high-frequency details를 보유하고 있는 것을 육안으로 확인가능하다. 이와 더불어, 잘못된 scene geometry 추정으로인해 생성된 blurry and foggy artifact들을 our models가 제거할 수 있는 능력을 보유하고 있음을 확인가능하다.
As shown in Fig.~\ref{fig:mainvisual1}, our models have more apparent textures and high-frequency details than IBRNet, GeoNeRF, GNT, Neuray, and MuRF. In addition, our models can remove blurry and foggy effects generated by incorrectly estimated scene geometry.

\noindent\textbf{Experiment on pixelSplat~\cite{charatan2023pixelsplat}.}
%우리는 최근 work로 소개된 3D gaussian splatting 기반의 generalizable rendeing model인 pixelSplat에서 본 방법론을 이용한 성능향상을 정량적, 정성적으로 입증한다. 본 실험을 위해, 우리 모델은 pixelSplat와 동일하게 ACID~\cite{infinite}, RealEstate10k~\cite{stereo} datasets으로 학습된다. Table~\ref{pixelsplat}에서 보이듯이, 우리는 두 개의 데이터셋에서 모두 기존 모델들 대비하여 가장 높은 성능을 가진다는 것을 확인할 수 있다. 또한, 모든 metrics에 대해서 기존 pixelSplat보다 ours를 사용했을때 성능향상이 있음을 확인가능하다. Fig.~\cite{fig:mainvisual}에서 보이듯이, 우리 모델은 pixelSplat가 합성하지 못하는 복잡한 object와 texutre를 artifact없이 완벽하게 synthesis하는 것을 정성적으로 확인가능하다.
We quantitatively and qualitatively demonstrate the performance improvement using our method in pixelSplat~\cite{charatan2023pixelsplat}, a generalizable rendeing model based on 3D gaussian splatting~\cite{kerbl20233d}. For this experiment, our model is trained on the same ACID~\cite{liu2021infinite} and RealEstate10k~\cite{zhou2018stereo} datasets as pixelSplat. As shown in Table~\ref{table:pixelsplat}, it has the highest performance compared to existing models in both datasets. In addition, there is a performance improvement when using ours compared to the results of pixelSplat. As shown in Fig.~\ref{fig:mainvisual}, it can be qualitatively confirmed that our model perfectly synthesizes complex objects and textures without artifacts.

\noindent\textbf{Novel view synthesis with per-scene finetuning.}
%per-scene optimization을 수행하는 최근 fast-training NeRF models과의 comparison을 위해, 우리는 G-NeRF models에 대해서 per-scene finetuning을 수행하고, 또한 finetuned G-NeRF models에 대해서 우리 모델이 image enhancement를 일으키는지에 대한 실험을 수행한다. Table 2에서와 같이, LLFF, DTU dataset에서 fast-training NeRFs(Instant-NGP, Tri-MipRF)보다 finetuned G-NeRFs(Neuray-ft, GNT-ft)의 성능이 더 높다. 게다가, 우리 모델이 부착된 G-NeRF models(Neuray-ft+Ours, GNT-ft+Ours)는 부착되지 않은 모델과 비교했을 때 most of metrics에 대해서 높은 성능향상을 일으킨다는 것을 확인할 수 있다. 
%
For comparison with recent novel view synthesis models, we perform per-scene finetuning on G-NeRF models and image enhancement on the finetuned G-NeRF models. As shown in Table~\ref{table:ft_llff}, G-NeRF models with ours (Neuray-ft~\cite{liu2022neural}+Ours, GNT-ft~\cite{wang2022attention}+Ours) are higher than NeRF-based models~\cite{mildenhall2021nerf,wizadwongsa2021nex} light-field based models~\cite{mildenhall2019local,suhail2022light} on Real Forward-Facing datasets. In addition, G-NeRF models with ours (Neuray-ft+Ours, GNT-ft+Ours) produce high performance improvements for most metrics than those without.
% \vspace{-10pt}

\subsection{Comparison with Image Enhancement Models}
\noindent\textbf{Experimental Setup.}
%이번 섹션에서, 우리는 기존 reference based 또는 frame based image enhancement models과의 성능비교를 진행한다. 우리는 SOTA models 중에서, single reference based models을 대표하여 C2-matching[]을 선정하고, multi reference based models을 대표하여 MRefSR[]을 선정한다. 우리는 또한 최근 video frame interpolation을 응용하여 NeRF에 적용한 frame based model인 NeRFLiX를 선정한다. 모든 비교모델들은 우리의 training dataset으로부터 training되며, 우리 모델을 포함한 모든 모델들의 G-NeRF base model은 Neuray로 통일된다.
In this section, we compare performance with existing reference-based or frame-based image enhancement models. Among state-of-the-art models, we select C2-matching~\cite{jiang2021robust} to represent single reference-based models and MRefSR~\cite{zhang2023lmr} to represent multi reference-based models. We also select NeRFLiX~\cite{zhou2023nerflix}, a frame-based model recently applied to NeRF by applying video frame interpolation. C2-matching and MRefSR are trained in our training dataset. Nerflix is fine-tuned on our training dataset using a pretrained model. For all models, including ours, G-NeRF base model is Neuray~\cite{liu2022neural}.

\noindent\textbf{Quantitative Analysis.}
% Table~\ref{}에서 보이듯이, 우리 모델은 모든 metrics, 그리고 모든 데이터셋에 대해서 C2-matching, MRefSR, NeRFLix보다 더 높은 성능을 가진다는 것을 확인할 수 있다. 따라서, 우리 모델은 기존 reference based 또는 frame based image enhancement model들과 비교했을 때 G-NeRF로부터 생성된 alpha point cloud 및 rendered image를 통해 geometry를 활용하여 정확한 reference based texture transfer를 수행함을 알 수 있다.
As shown in Table~\ref{table:refsr}, our model outperforms C2-matching, MRefSR, and NeRFLiX for all metrics and datasets. 
% In the DTU dataset, ours has a 28.72 PSNR which is the highest value. In the Synthetic NeRF dataset, ours has a 0.17 higher PSNR than NeRFLiX and is the highest-valued model. In the Real Forward-Facing datset, ours has a 25.82 PSNR which is the highest value. In evaluation metrics SSIM and LPIPS other than PSNR, our model also shows superiority with respect to other multi-reference-based image enhancement studies.
We also compare the inference time, model trainable parameters, and the number of reference view inputs.
As shown in Table~\ref{table:enhance_speed}, C2-Matching has the least parameters and uses only one reference view, so it has the fastest inference time but the lowest PSNR value.
On the other hand, compared to other models that use multi-reference views (MRefSR, NeRFLiX), our model has the smallest number of parameters at 9.1 million and the fastest inference time for all datasets. Additionally, our model has the highest PSNR value compared to all models.
%This comparison shows that our model transfers textures better than existing reference-based or frame-based image enhancement models by utilizing geometry through the alpha point cloud.
% , our model performs accurate reference-based texture transfer

\begin{table}[]
% \vspace{-20pt}
\centering
\caption{Novel view synthesis results with various reference-based image enhancement models. For a fair comparison, we use Neuray as the G-NeRF baseline for all models.
%\textbf{Bold} indicates the best, and \underline{underline} indicates the second best results.
}
% \vspace{-8pt}
\resizebox{0.93\textwidth}{!}{%
\setlength{\tabcolsep}{3pt}
\begin{tabular}{c||c|c|c||c|c|c||c|c|c}
\Xhline{3\arrayrulewidth}
\multirow{2}{*}{Model} &  \multicolumn{3}{c||}{DTU}    &   \multicolumn{3}{c||}{Synthetic NeRF}   &      \multicolumn{3}{c}{Real Forward-Facing}       \\
\cline{2-10}
           & PSNR($\uparrow$)    & SSIM($\uparrow$)    & LPIPS($\downarrow$)  & PSNR($\uparrow$)    & SSIM($\uparrow$)    & LPIPS($\downarrow$)  & PSNR($\uparrow$)    & SSIM($\uparrow$)    & LPIPS($\downarrow$) \\
\Xhline{3\arrayrulewidth}
C2-Matching~\cite{jiang2021robust}        &  28.38    &  \underline{0.910}    &   0.105 &  28.49    & \underline{0.928}    &  0.068     &   25.55   &   0.839   &    0.154   \\
\hline
MRefSR~\cite{zhang2023lmr}       &  28.52    & 0.903     &   0.104    &    28.64  &    0.917  &     0.071  &   25.64   &  \underline{0.844}    &    0.144   \\
\hline
NeRFLiX~\cite{zhou2023nerflix}  &   \underline{28.54}   &  0.906    &  \underline{0.103}     &  \underline{28.79}    &   0.926   &  \textbf{0.062}     &  \underline{25.65}    &   \underline{0.844}   &  \underline{0.143} \\     
\hline
Ours        &    \textbf{28.72}  &  \textbf{0.920}    &  \textbf{0.101}    &   \textbf{28.96}   &   \textbf{0.936}   &   \textbf{0.062}    &   \textbf{25.82}   &  \textbf{0.850}    & \textbf{0.142} \\
\Xhline{3\arrayrulewidth}
\end{tabular}
}
% \vspace{-10pt}
\label{table:refsr}
\end{table}
\begin{figure*}[t]
    \centering
 %   \vspace{3pt}
    \includegraphics[width=0.9\linewidth]{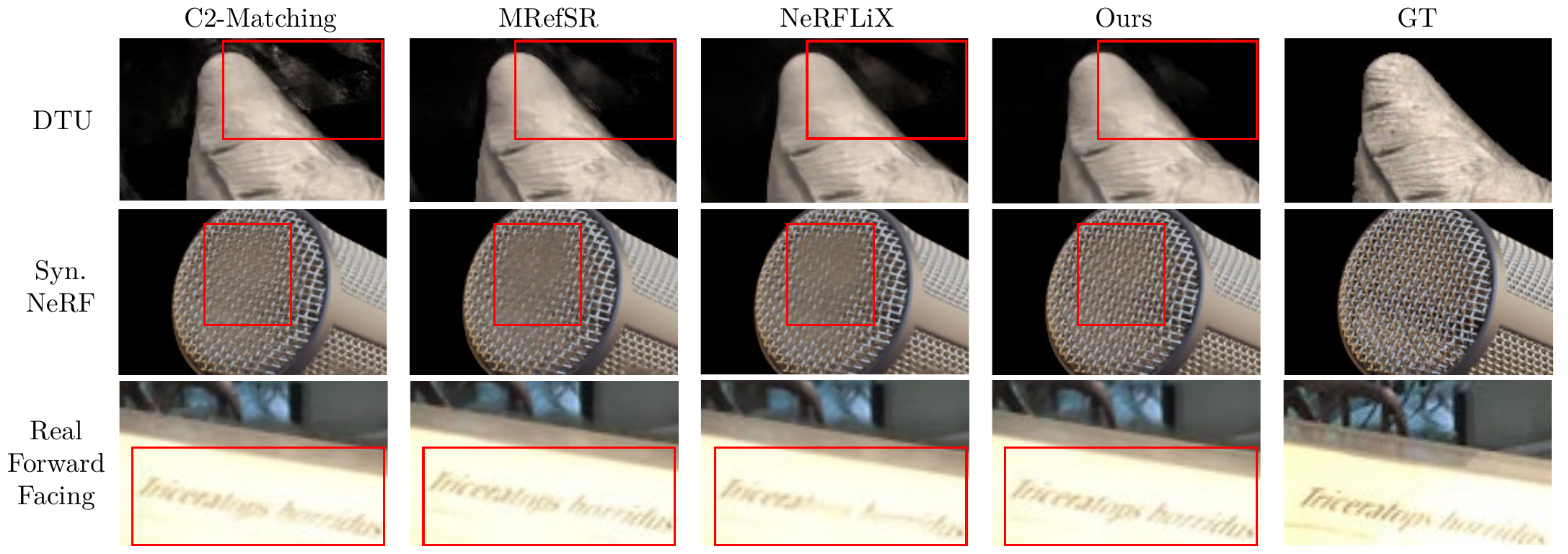}
    % \vspace{-5pt}
    \caption{Qualitative comparisons of reference-based image enhancement on DTU, Real Forward-Facing, and Synthetic NeRF dsatasets.}
     % \vspace{-17pt}
    \label{fig:ref}
\end{figure*}
% \vspace{70pt}

\noindent\textbf{Qualitative Comparison and Analysis.}
%Fig.~\ref{}에서 보이듯이, 제안된 모델은 image enhancement 과정에서 탁월한 선명성과 텍스쳐 세부성을 제공한다. 더불어, 안개 효과를 효과적으로 제거하여 시각적 환경을 개선하는 역할을 수행한다. 우리 모델은 선명한 텍스쳐를 생성함으로써 이미지의 고품질 표현을 도모하며, 동시에 안개 효과의 소거를 통해 시각적 잡음을 최소화한다. 
As shown in Fig.~\ref{fig:ref}, our model provides image clarity and texture details in comparison to other models. 
Fine textures such as mic patterns and letters are well transferred from source images.
In addition, our model effectively removes fog effects and generates clear textures, thereby promoting high-quality representation of images and minimizing noise.

% \vspace{-25pt}

\subsection{Ablation Studies}
% \vspace{-5pt}
%이번 섹션에서, 우리는 세 가지에 대한 ablation studies를 진행한다. 첫 번째로, offset 추정 시 alpha 및 correspondence의 존재에 따른 novel-view synthesis 성능을 비교한다. 두 번째로, Feature aggregation과정에서 모델 architecture에 따른 성능을 비교한다. 마지막으로, reference image 수에 따른 성능비교를 통해 모델이 reference image에 의존하는지에 대한 여부를 판단한다.
In this section, we conduct two ablation studies. 
First, we analyze the influence of alpha and correlation values on the RayDCN.
% Second, we compare performance according to model architecture during the feature aggregation process. 
Second, we examine the effect of varying model architectures on performance in the feature aggregation process.
% Finally, we determine whether the model depends on the reference image by comparing performance according to the number of reference images.

\noindent\textbf{RayDCN.}
%우리 모델은 DCN의 input으로 사용되는 offset을 추정하기 위해 alpha와 correspondence values를 이용한다. 우리는 alpha와 correspondence의 필요성을 파악하기 위한 ablation study를 진행한다. Table~\ref{}에서 보이듯이, correspondace만 오직 사용할 경우 alpha만 사용할 때보다 더 낮은 성능을 가진다. 또한, 본 논문에서 제안한 offset estimation network를 통해 두 values를 모두 이용하면, 하나만 사용할때와 비교했을 때 가장 높은 성능을 가진다는 것을 확인할 수 있다.
% Our model uses alpha and correspondence values to estimate the offset used as input to the DCN. We conduct an ablation study to determine the need for alpha and correspondence. As shown in Table~\ref{table:offset_ablation}, when only correspondace is used, performance is lower than when only alpha is used. In addition, it can be confirmed that using both values through the offset estimation network proposed in this paper has the highest performance compared to using only one.
% Our model uses alpha and correspondence values to estimate the offset used for the input of DCN. We conduct an ablation study to determine the need for alpha and correspondence.
% As shown in Table~\ref{table:offset_ablation}, performance is lower when only correspondence is used than when only alpha is used. In addition, it can be confirmed that using both values through the offset estimation network proposed in this paper has the highest performance compared to using only one.
% Table~\ref{table:offset_ablation} reveals that performance is diminished when only correspondence is utilized, as opposed to using only alpha. Furthermore, it is evident that employing both values through the offset estimation as RayDCN yields the highest performance compared to using only one.
We perform an ablation study to determine the effectiveness of geometry proxy $X^{alpha}$ and image-based correlation $Corr$ as inputs to RayDCN. 
We categorize the input types for RayDCN into three cases: the first utilizes only $Corr$, the second employs only $alpha$, and the third incorporates both inputs. The first case represents the original DCN, while the third case corresponds to the proposed complete RayDCN.
Table~\ref{table:offset_ablation} shows that performance diminishes when alpha and correlation are used singly; the proposed RayDCN, which utilizes alpha and correlation, shows substantial performance improvement.

\begin{figure}[h]
    \centering
    \includegraphics[width=0.95\linewidth]{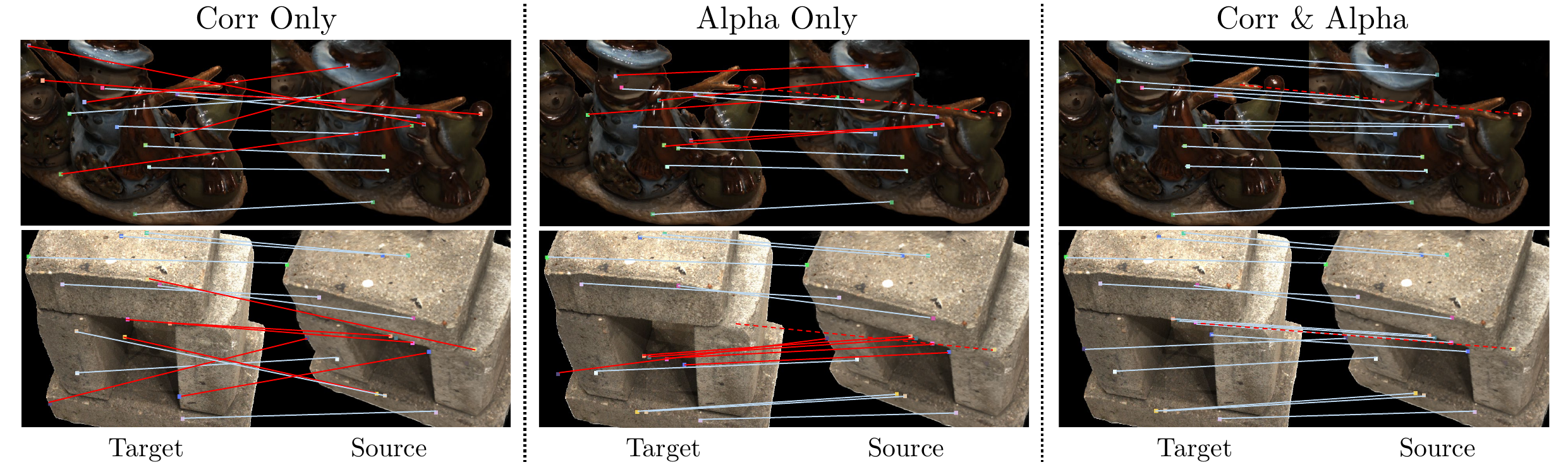}
    % \vspace{-8pt}
    % \caption{Offset estimation outputs on the DTU dataset. {\color{blue}{Blue}} lines denote instances of correct offset estimation. {\color{red}{Red}} lines denote instances of incorrect offset estimation, while dotted lines signify instances of missing offsets where the projected point on the source view is out of the source image.}
    \caption{Offset estimation results on the DTU dataset: {\color{blue}{Blue}} lines indicate correct estimates, {\color{red}{Red}} lines indicate incorrect estimates, and dotted lines signify missing offsets.}
    % where the projected point is outside the source image.}    

    % \vspace{-20pt}
    \label{fig:offset}
\end{figure}

For qualitative comparison, Fig. ~\ref{fig:offset} present offset estimation visualizations under varied input conditions. For a randomly sampled single coordinate $p_0$ in the rendered image, offset estimation yields a projected point $p'=p+\mathbf{H}(p)$ on the source image. 
As a result, employing complete RayDCN(Corr \& alpha) ensures that $p'$ and $p$ correspond to the same 3D point, effectively constraining the position of $p'$ within the epipolar line and demonstrating improved accuracy in estimating 3D point on the surface of an object.

\clearpage

\begin{table}[t]
\centering
\caption{Ablation study of different input types for RayDCN with $Corr$ and $Alpha$ as inputs. %RayDCN with $Corr$ only is the same as the original deformable convolution. 
\textbf{Bold} indicates the best, and \underline{underline} indicates the second best results.}
% \vspace{-6pt}
\resizebox{0.93\textwidth}{!}{%
\setlength{\tabcolsep}{3pt}
\begin{tabular}{c||c|c|c||c|c|c||c|c|c}
\Xhline{3\arrayrulewidth}
\multirow{2}{*}{Input Type} &  \multicolumn{3}{c||}{DTU}    &   \multicolumn{3}{c||}{Synthetic NeRF}   &      \multicolumn{3}{c}{Real Forward-Facing}       \\
\cline{2-10}
           & PSNR($\uparrow$)    & SSIM($\uparrow$)    & LPIPS($\downarrow$)  & PSNR($\uparrow$)    & SSIM($\uparrow$)    & LPIPS($\downarrow$)  & PSNR($\uparrow$)    & SSIM($\uparrow$)    & LPIPS($\downarrow$) \\
\Xhline{3\arrayrulewidth}
$Corr$ Only       &  28.56    &   0.916   &   \underline{0.105}    &  28.68    &   \textbf{0.936}   &  \underline{0.065}     &  25.67    &    0.832  &    0.156   \\
\hline
$Alpha$ Only      &   \underline{28.57}   &   \underline{0.917}   &  0.109     &    \underline{28.77}  &   0.937   &  \underline{0.065}     &   \underline{25.73}   &   \underline{0.846}   &   \underline{0.155}    \\
\hline
$Corr$ \& $Alpha$  &   \textbf{28.72}   &   \textbf{0.920}   &   \textbf{0.101}    &   \textbf{28.96}   &   \textbf{0.936}   &   \textbf{0.062}    &  \textbf{25.82}    & \textbf{0.850}     &\textbf{0.142} \\     
\Xhline{3\arrayrulewidth}
\end{tabular}
}
% \vspace{-6pt}
\label{table:offset_ablation}
\end{table} 
\begin{table}[t]
% \vspace{-10pt}
\centering
\caption{Ablation study of the aggregation types. \textbf{Bold} indicates the best, and \underline{underline} indicates the second best results.}
% \vspace{-10pt}
\resizebox{0.93\textwidth}{!}{%
\setlength{\tabcolsep}{3pt}
\begin{tabular}{c||c|c|c||c|c|c||c|c|c}
\Xhline{3\arrayrulewidth}
\multirow{2}{*}{Agg. Type} &  \multicolumn{3}{c||}{DTU}    &   \multicolumn{3}{c||}{Synthetic NeRF}   &      \multicolumn{3}{c}{Real Forward-Facing}       \\
\cline{2-10}
           & PSNR($\uparrow$)    & SSIM($\uparrow$)    & LPIPS($\downarrow$)  & PSNR($\uparrow$)    & SSIM($\uparrow$)    & LPIPS($\downarrow$)  & PSNR($\uparrow$)    & SSIM($\uparrow$)    & LPIPS($\downarrow$) \\
\Xhline{3\arrayrulewidth}
MLP  &  28.54    &  0.915    &  0.110    &   28.61    & 0.934     &   0.068   &  25.63     & 0.842     &  \underline{0.153}           \\
\hline
Multi-Head Attention      &  \underline{28.58}    &   \underline{0.916}   & \underline{0.106}      &  \underline{28.69}    &    \underline{0.935}  &    \underline{0.066}   &    \underline{25.67}  &  \underline{0.844}    &   0.154    \\
\hline
TPFormer  &  \textbf{28.72}    &   \textbf{0.920}   &    \textbf{0.101}   &  \textbf{28.96}    &   \textbf{0.936}   &    \textbf{0.062}   &  \textbf{25.82}    &   \textbf{0.850}   & \textbf{0.142}\\     
\Xhline{3\arrayrulewidth}
\end{tabular}
}
% \vspace{-18pt}
\label{table:fusion_ablation}
\end{table} 

\noindent\textbf{TPFormer.}
%우리는 source image feature aggregation을 위한 model architecture에 대해서 ablation study를 진행한다. 첫 번째는 IBRNet, Neuray에서 사용한 MLP layer 구조이고, 두 번째는 GNT에서 제안한 Multi-Head Attention(MHA)이다. 마지막으로 세번째는 본 논문에서 제안한 ref-transformer이다.  Table~\ref{}에서 보이듯이, 세 데이터셋 모두에 대해 세 개의 metrics에 대해서 본 논문이 제안한 ref-transformer module이 가장 좋은 성능을 보이는 것을 확인할 수 있다.
We conduct an ablation study on model architecture for source image feature aggregation. The first is the MLP layer structure used in IBRNet~\cite{wang2021ibrnet} and Neuray~\cite{liu2022neural}, the second is the Multi-Head Attention (MHA) proposed by GNT~\cite{wang2022attention}, and the third is TPFormer proposed in this paper. As shown in Table~\ref{table:fusion_ablation}, TPFormer shows the best performance for the three metrics for all three datasets.
Specifically, TPFormer outperforms the MLP layer, as well as MHA. The superior performance indicates the effectiveness of TPFormer in aggregating source image features, making it a promising choice for enhancing model architecture in tasks related to feature fusion and aggregation. 
\section{Conclusion}
% \vspace{-5pt}
In this paper, we propose a geometry-driven multi-reference texture transfer network, called GMT, to enhance generalizable neural rendering. We also propose novel modules, RayDCN and TPFormer, for aggregating the local texture of source images using a rendered image and an alpha point cloud generated from the volume densities of G-NeRF. We demonstrate consistent improvement for various datasets using most G-NeRF as based models.

\noindent\textbf{Acknowledgement}

\noindent This work was supported by the National Research Foundation of Korea(NRF) grant funded by the Korea government(MSIT) (NRF2022R1A2B5B03002636).

% \clearpage\mbox{}Page \thepage\ of the manuscript.
% \clearpage\mbox{}Page \thepage\ of the manuscript.
% \clearpage\mbox{}Page \thepage\ of the manuscript.
% \clearpage\mbox{}Page \thepage\ of the manuscript.
% \clearpage\mbox{}Page \thepage\ of the manuscript. This is the last page.
% References should start immediately after the main text, but can continue past p.\ 14 if needed.
% \clearpage  % TODO REVIEW/FINAL: This \clearpage needs to be removed from both review and camera-ready versions.

\bibliographystyle{splncs04}
% \bibliography{main}

\end{document}